\documentclass[nohyperref]{article}

\usepackage{microtype}
\usepackage{graphicx}
\usepackage{booktabs} % for professional tables

\usepackage{hyperref}

\usepackage[accepted]{icml2022}

\usepackage{amsmath}
\usepackage{amssymb}
\usepackage{mathtools}
\usepackage{amsthm}

\usepackage{subcaption}
\usepackage{csquotes}
\usepackage{multirow}
\usepackage{cuted}
\usepackage{pifont}
\usepackage{algorithmic}
\usepackage{algorithm}
\usepackage{bbm}
\usepackage{color}
\usepackage{contour}

\usepackage[capitalize,noabbrev]{cleveref}

\theoremstyle{plain}
\newtheorem{definition}{Definition}[section]
\newtheorem{theorem}{Theorem}[section]

\newtheorem{lemma}{Lemma}[section]

\newcommand\negative{\textrm{-}}

\newcommand\M{\textbf{M}}
\newcommand\U{\textbf{U}}
\newcommand\V{\textbf{V}}

\newcommand\F{\textbf{F}}
\newcommand\I{\textbf{I}}
\newcommand\W{\textbf{w}}
\newcommand\Q{Q}
\newcommand\Prob{\mathbb{P}}

\newcommand\1{\mathbbm{1}}
\newcommand\eps{\boldsymbol{\epsilon}}
\newcommand\Eps{\textbf{E}}

\newcommand\ul{\textbf{u}}
\newcommand\X{\textbf{x}}
\newcommand\T{\textbf{t}}
\newcommand\Y{\textbf{y}}
\newcommand\dpa{\partial}

\contourlength{0.5pt}

\newcommand{\cmark}{\contour{green}{\normalsize \color{black}\ding{51}}}
\newcommand{\xmark}{\contour{red}{\normalsize \color{black}\ding{55}}}

\newcommand\numberthis{\addtocounter{equation}{1}\tag{\theequation}}

\icmltitlerunning{SBI: A Simulation-Based Test of Identifiability for Bayesian Causal Inference}

\begin{document}

\twocolumn[
\icmltitle{SBI: A Simulation-Based Test of Identifiability \\for Bayesian Causal Inference}




\begin{icmlauthorlist}

\icmlauthor{Sam Witty}{Basis,UMass}
\icmlauthor{David Jensen}{UMass}
\icmlauthor{Vikash Mansinghka}{MIT}

\end{icmlauthorlist}

\icmlaffiliation{UMass}{College of Information and Computer Sciences, University of Massachusetts, Amherst, United States}
\icmlaffiliation{MIT}{Department of Brain and Cognitive Sciences, Massachusetts Institute of Technology, Cambridge, United States}
\icmlaffiliation{Basis}{Basis Research Institute, New York, United States}

\icmlcorrespondingauthor{Sam Witty}{sam@basis.ai}

\icmlkeywords{Machine Learning, ICML}

\vskip 0.3in
]



\printAffiliationsAndNotice{}  % leave blank if no need to mention equal contribution

% \begin{abstract}
% This paper introduces a procedure for testing the identifiability of queries in causal probabilistic programs. Although the do-calculus is sound and complete given a causal graph, many practical assumptions cannot be expressed in terms of graph structure alone, such as the assumptions required by instrumental variable designs, regression discontinuity designs, and within-subjects designs. We present simulation-based identifiability (SBI), a fully automated identification test based on a particle optimization scheme with simulated observations. SBI allows users to determine whether queries of custom causal models are identifiable without custom mathematical analysis. We analyze SBI's soundness for a broad class of differentiable, finite-dimensional causal probabilistic programs with bounded effects. Finally, we provide an implementation of SBI using stochastic gradient descent that applies to a broad class of causal probabilistic programs, and show empirically that it agrees with known graph-based and quasi-experimental design identification results, including those using Gaussian processes.
% \end{abstract}

\begin{abstract}
A growing family of approaches to causal inference rely on Bayesian formulations of assumptions that go beyond causal graph structure. For example, Bayesian approaches have been developed for analyzing instrumental variable designs, regression discontinuity designs, and within-subjects designs. This paper introduces simulation-based identifiability (SBI), a procedure for testing the identifiability of queries in Bayesian causal inference approaches that are implemented as probabilistic programs. SBI complements analytical approaches to identifiability, leveraging a particle-based optimization scheme on simulated data to determine identifiability for analytically intractable models. We analyze SBI's soundness for a broad class of differentiable, finite-dimensional probabilistic programs with bounded effects. Finally, we provide an implementation of SBI using stochastic gradient descent, and show empirically that it agrees with known identification results on a suite of graph-based and quasi-experimental design benchmarks, including those using Gaussian processes.
\end{abstract}

\begin{figure*}[t!]
\centering
\captionsetup[subfloat]{aboveskip=-0.4cm}
    \begin{subfigure}[b]{0.18\textwidth}
        \centering
        \includegraphics[width=0.7\textwidth]{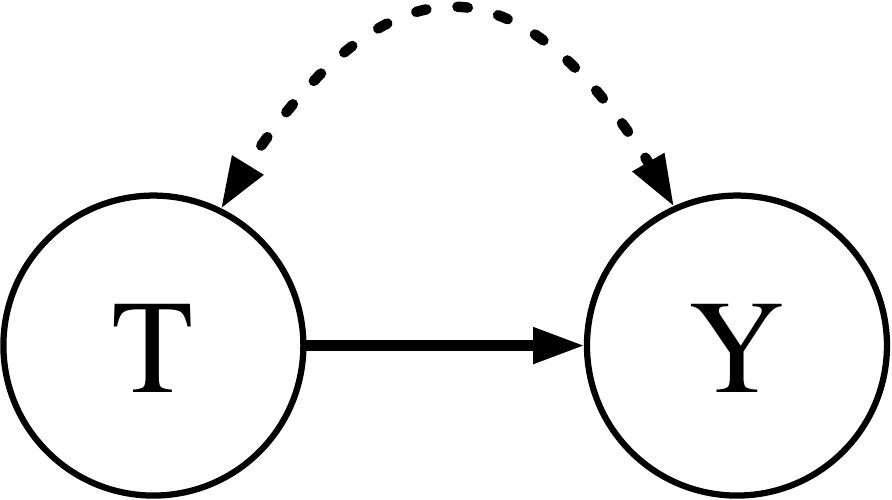}
        \vspace{-0.2cm}
        \footnotesize{
        \begin{align*}
        T_i &= f_T(U_i, \epsilon_{T_i}; \theta) \\
        Y_i &= f_Y(T_i, U_i, \epsilon_{Y_i}; \theta)
        \end{align*}
        }
        \vspace{-1.5cm}
    \end{subfigure}
    ~
    \begin{subfigure}[c]{0.25\textwidth}
        \centering
        \includegraphics[width=\textwidth]{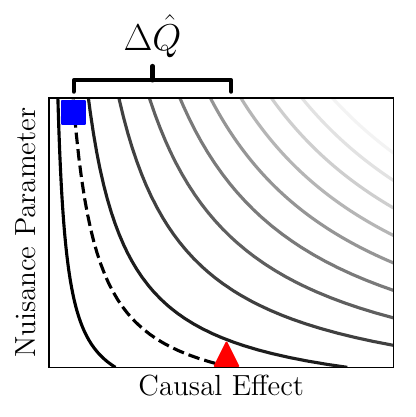}
        \label{fig:conf_like}
    \end{subfigure}
    ~
    \begin{subfigure}[c]{0.25\textwidth}
        \centering
        \includegraphics[width=\textwidth]{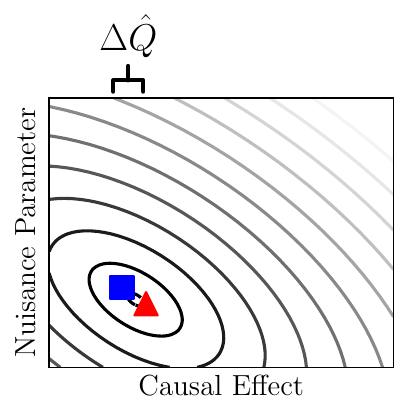}
        \vspace{0.08cm}
    \end{subfigure}
    ~
    \begin{subfigure}[b]{0.18\textwidth}
        \centering
        \includegraphics[width=\textwidth]{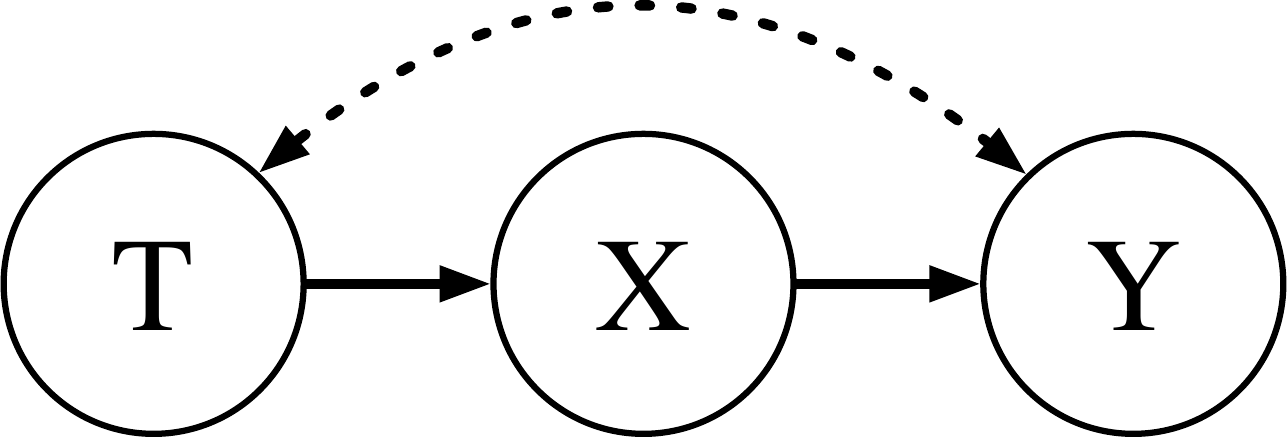}
        \vspace{-0.5cm}
        \footnotesize{
        \begin{align*}
        T_i &= f_T(U_i, \epsilon_{T_i}; \theta) \\
        X_i &= f_X(T_i, \epsilon_{X_i}; \theta)\\
        Y_i &= f_Y(X_i, U_i, \epsilon_{Y_i}; \theta)
        \end{align*}
        }
        \vspace{-1.7cm}
    \end{subfigure}
    
    % \hspace{.25\textwidth}
    
    \subfloat[\label{fig:not_id_contour} Non-identifiable causal model and likelihood]{\hspace{.5\textwidth}}
    \subfloat[\label{fig:id_contour} Identifiable causal model and likelihood]{\hspace{.5\textwidth}}
    % \vspace{1cm}
    % \vspace{1cm}
    
    \caption{\footnotesize{\textbf{Overview of simulation-based identiability.} Simulation-based identifiability (SBI) recasts causal identifiability as an optimization problem that seeks to maximize the data likelihood \textit{and} the distance, $\Delta \hat{Q}$, between the effect estimates induced by two sets of parameters, $\theta^{(1)}$ and $\theta^{(2)}$. When causal effects are not identifiable (a) SBI discovers maximum likelihood parameters (blue and red) that estimate different causal effects. When causal effects are identifiable (b) the two models converge to the same effect estimates.
    }}
    \label{fig:intuition}
\end{figure*}
\raggedbottom

\section{Introduction}

Drawing causal conclusions from data requires assumptions about underlying causal mechanisms~\cite{pearl2009causality}. Consequently, it is important to determine when these assumptions are sufficient to answer a causal query, i.e. whether the query is \textit{identifiable}. Existing computational methods, such as the do-calculus, can rigorously determine identifiability from graph structure alone~\cite{huang2006pearl, pearl1995causal}, however, graph structure alone can be incomplete in some cases. For example, instrumental variable designs require an assumption of monotonicity or linearity~\cite{cragg1993testing}, within-subjects designs require an assumption that latent confounders are shared across units~\cite{gelman2006multilevel, loftus1994using}, and regression discontinuity designs violate positivity, an assumption required by the do-calculus~\cite{lee2010regression}.
% Unfortunately, the do-calculus does not provide insight in these settings and other techniques are necessary.
%The do-calculus is inconclusive in these settings because it ignores the restrictions on structural functions that make the causal effects identifiable. 

A growing body of causal inference research employs assumptions that go beyond graph structure. For example, researchers in causal machine learning~\cite{athey2006identification, hartford2017deep}, and in hierarchical probabilistic modeling approaches to causal inference \cite{branson2019nonparametric, louizos2017causal, tran2018implicit, witty2020causal}, have achieved promising results. Some of these techniques can be expressed as priors over structural causal models, and implemented as probabilistic programs.

Unfortunately, it is difficult to apply either analytical or graphical techniques to determine the identifiability of complex Bayesian approaches to causal inference. As a result, these approaches can produce inaccurate effect estimates even with infinite data~\cite{d2019multi, rissanen2021critical}. This paper introduces new automated techniques that can improve the rigor of causal inferences by providing simulation-based tests of identifiability (SBI). 

SBI is compatible with any prior over structural causal models that: (i) can be used to sample data; and (ii) induces a differentiable likelihood function. The key innovation is to reduce causal identification to an optimization procedure that maximizes the likelihood of two sets of parameters while also maximizing the distance between their causal effect estimates. If the optimal solution is two sets of parameters that agree on effect estimates, then the effect is identifiable. See Figure~\ref{fig:intuition} for intuition.

%Historically, practitioners with non-graphical assumptions have had to choose between a small number of custom recipes scattered throughout the literature with analytic solutions, which often make strong parametric assumptions. Instead, we would like to build on recent advances in causal machine learning while leveraging custom non-graphical assumptions. In fact, a large and growing community at the intersection of probabilistic machine learning and causal inference has developed models to serve exactly this purpose, and have achieved promising empirical performance~. Unfortunately, these kinds of models, expressible as , are not covered by existing automated identification techniques, and may 

% This paper proposes an alternative: (i) express assumptions as causal probabilistic programs, inducing a joint distribution over observations and counterfactuals; and (ii) automatically search for models that are likelihood equivalent, but that produce different effect estimates.

%We introduce simulation-based identifiability (SBI), a flexible and automated test of identifiability for queries in causal probabilistic programs, i.e. Bayesian priors over parametric structural causal models. 

In Section~\ref{sec:Optimization}, we prove that SBI is asymptotically sound and complete, assuming certain (strong) regularity conditions. In Section~\ref{sec:Experiments}, we show that SBI is broadly applicable by presenting a suite of compatible benchmarks reflecting common graph-based and quasi-experimental designs. We show empirically that SBI correctly determines whether average treatment effects are identifiabile for all fourteen benchmarks. Finally, we use SBI to extract quantitive insight about Gaussian process regression discontinuity designs.

\subsection{Related Work}

Our work is not the first to automate identification for causal inference. Symbolic methods for observational~\cite{huang2006pearl, pearl1995causal} and experimental~\cite{lee2020general} data determine whether queries are nonparametrically identifiable using graph structure alone. Similar methods have been developed for linear models~\cite{bollen2005structural, kumor2019efficient}. When applicable, symbolic methods like the do-calculus should be the de-facto choice, as they have strong theoretical gaurantees, are computationally efficient, and require minimal ancillary assumptions. However, these approaches are inconclusive for more flexible parameterizations, such as those using Gaussian processes, or models employing non-graphical assumptions, such as within-subjects designs. These methods (and SBI) do not attempt to test whether a set of assumptions are satisfied given a particular dataset. Instead, they test whether assumptions are sufficient to uniquely determine a causal effect from (yet unseen) data.

Similar approaches for determining identifiability have been developed in other fields, such as neuroscience~\cite{valdes2011effective} and dynamical systems~\cite{raue2009structural}, by searching for likelihood equivalent parameters using gradient-based search. SBI differs from these approaches in two important ways. First, SBI uses a particle-based objective function to search for likelihood equivalent models globally, rather than locally near a single maximum likelihood solution. Second, SBI's objective function searches for models that estimate different causal effects, not only different parameters. This distinction means that SBI can correctly determine identifiability even when effects are composed of many parameters (e.g. see Section~\ref{sec:gp_example}). It is well known that queries can be identified even in settings where individual parameters cannot~\cite{pearl2009causality}. Optimization techniques have been used to bound counterfactual queries~\cite{balke1994counterfactual, tian2000probabilities, zhang2021partial} or for neural-causal models~\cite{xia2021causal}, but do not support user-specified parametric assumptions.

Bayesian priors over parametric structural causal models can be implemented in probabilistic programming languages~\cite{goodman2008church, mansinghka2014venture}, which provide a syntax for expressing probabilistic models as code. Many of these languages support automatic differentiation and gradient-based optimization~\cite{bingham2018pyro, carpenter2017stan, cusumano-towner2019gen, dillon2017tensorflow}, providing the necessary utilities for our optimization-based approach. While some languages contain an explicit representation of interventions~\cite{bingham2018pyro, perov2020multiverse, tavares2019counterfactual, witty2020bayesian}, none currently address causal identifiabilty.

\begin{table*}[t!]
\centering
    \footnotesize
  \begin{tabular}{p{0.9in} p{4.3in} c}
    \toprule
      \textbf{Design} & \textbf{Description} & \textbf{Source}\\
      \midrule
      \midrule
      Unconfounded & No latent variables influence both treatment, $T$, and outcome, $Y$. &  \cite{pearl1995causal}\\
      \midrule
      Confounded & A latent confounder, $U$, influences both $T$ and $Y$. & \cite{pearl1995causal}\\
      \midrule
      Backdoor & An observed confounder, $X$, influences $T$ and $Y$. & \cite{pearl1995causal}\\
      \midrule
      Frontdoor & $U$ influences $T$ and $Y$, but does not influence a mediator, $X$. &  \cite{pearl1995causal}\\
      \midrule
      Instrumental \;\;\;\;variable & $U$ influences $T$ and $Y$. An observed instrument, $I$, influences $T$, does not influence $Y$ except through $T$, and is not influenced by $U$. & \cite{angrist1996identification}\\
      \midrule
      Within subjects & Each instance of $U$ influences multiple instances of $T$ and $Y$. & \cite{draper1995inference}\\
      \midrule
      Regression \newline discontinuity & An observed confounder, $X$, influences $T$ and $Y$. $T$ is fully determined by $X$ being above or below a known threshold. & \cite{rubin1977assignment}\\
    \bottomrule
  \end{tabular}
  \caption{\footnotesize{
  \textbf{Description of quasi-experimental designs benchmarks.} Of these seven standard causal designs, intrumental variable, within subjects, and regression discontinuity designs require assumptions that go beyond graph-structure. Parameterized versions of all seven designs can be represented as probabilistic programs, and can thus be tested using simulation-based identifiability.}}
  \label{tab:design_desc}
\end{table*}

\section{Preliminaries}

In this section we describe the necessary background on causal inference, and the Bayesian approach that SBI supports. Throughout this paper we use lowercase bold letters to denote arrays of random variables and uppercase letters with subscripts to denote their elements, e.g. $\ul_1 = [U_{1,1},.., U_{1, n}]$. We use $Q$ interchangeably depending on context to denote either a causal query, e.g. sample average treatment effect, or the random variable induced by applying $Q$ to its random inputs.

\begin{definition}
    \textbf{Structural Causal Models.} A structural causal model (SCM) is a four-tuple $\M = \langle \V, \Eps, \U, \F \rangle$, where: $\V = \{\T, \Y, \X_1, ..., \X_d\}$ is a set of observed random variables, $\Eps = \{\eps_\T, \eps_\Y, \eps_{\X_1}, ..., \eps_{\X_d}\}$ is a set of independent exogenous latent variables, $\U = \{\ul_1, ..., \ul_{d'}\}$, is a set of latent confounders, and $\F = \{f_T, f_Y, f_{X_1}, ..., f_{X_d}\}$ is a set of deterministic functions. Each observed variable is assigned according to its corresponding structural function, e.g. $Y_i = f_Y(T_i, U_{1, i}, \eps_{Y_i})$. We use $Pa(\cdot)$ as shorthand for structural function arguments, e.g. $Pa(Y_i) = \{T_i, U_{1, i}, \eps_{Y_i}\}$. By construction, each $\eps \in \Eps$ is an argument of exactly one structural function, and each $\ul \in \U$ is an argument of at least two structural functions. To represent an intervention, $do(T_i = t')$, we replace the expression $T_i = f_T(Pa(T_i))$ with the expression $T_i = t'$ and leave all other structural functions unchanged. We use $\V'$ to denote the set of counterfactual random variables $\{\Y', \X'_1, ..., \X'_d\}$ induced by an intervention $do(T_i = t')$ for all $i$ in $1, ..., n$.\footnote{Here, we slightly modify the definition in~\cite{pearl2009causality} to distinguish between confounders and exogenous noise, and to clarify that we only consider interventions on a single variable $\T$ and queries on a single variable $\Y$.}
\end{definition}

% In practice, we rarely have access to the SCM, nor can we hope to learn one exactly from data. Instead, we make assumptions that partially restrict the class of SCMs a priori, and then ask whether SCMs that are consistent with observed data within that class also produce consistent effects~\cite{bareinboim2020pearl}. 
% Assuming a causal graph restricts the arguments of each structural function, but places no restrictions on functional form, nor on the distribution of confounders or exogenous noise. ~\cite{huang2006pearl, pearl1995causal}. 

% In many realistic scenarios, however, we may wish to restrict the class of SCMs further, including restrictions on functional form. For example, in a sharp \textit{regression discontinuity design}~\cite{lee2010regression} we know that treatment is fully determined by a covariate being above or below a known threshold $a$, i.e. $T_i = \1_{X_{1, i} > a}$, where $\1$ is the indicator function. In these and similar settings, 
A growing body of literature~\cite{louizos2017causal, perov2020multiverse, tavares2019counterfactual, tran2018implicit, witty2020bayesian, witty2020causal} has represented causal assumptions as a prior distribution over parametric SCMs, $p(\Eps, \F, \U)$. Here, practitioners do not claim that probability subsumes causal inference~\cite{pearl2001bayesianism}, but instead that it represents uncertain belief over a class of (still causal) SCMs. This prior serves a conceptually similar role to a causal graph, restricting the space of SCMs a-priori~\cite{bareinboim2020pearl}.

% This approach does not claim that probability theory subsumes causal inference as discussed in~\cite{pearl2001bayesianism}, only that it represents our uncertain belief over a class of (still causal) SCMs. In other words, the support of the prior plays the same conceptual role as a causal graph, albiet with stronger restrictions and in a different form.

It is often convenient to reason about such priors in terms of observed endogenous variables, $p(\V, \F, \U) = p(\V|\F, \U) p(\F, \U)$, marginalizing out exogenous noise. In this paper we assume that this \textit{pushforward}, or change of variables, is tractable. For example, for the linear parameterization $Y_i = \beta T_i + \alpha U_i + \gamma \epsilon_{Y_i}$, $\epsilon_{Y_i} \sim \mathcal{N}(0, 1)$, the conditional density $p(Y_i|f_Y, U_i)$ is given by $\mathcal{N}(Y_i; \beta T_i + \alpha U_i, \gamma)$.

% Given a distribution over exogenous noise, SCMs implicitly induce a joint density over observed random variables, $p(\V| \F, \U)$. We express this joint density conditional on structural functions, $\F$, and latent confounders, $\U$, for reasons that will become clear in Section~\ref{sec:Bayesian} when we discuss the Bayesian approach. Here, $\delta$ is the dirac-delta function.

% \begin{equation}
%     p(\V|\F, \U) = \prod_{i=1}^n \prod_{V \in \V} \int \delta(V_i - f_V(Pa(V_i)) \; dp(\eps_{V_i}) \label{eq:likelihood}
% \end{equation}

% While Equation~\ref{eq:likelihood} is not tractable in general, many structural functions lend themselves to closed-form conditional distributions.  

% Bayesian approach to causal inference involves placing a prior over structural functions and latent confounders, $p(\F, \U)$. 

Practitioners are rarely interested in counterfactual outcomes directly, and instead are interested in some causal query, $Q(\Y, \Y')$, such as the sample average treatment effect, $Q = \sum_{i=1}^n (Y'_{i} - Y_i)/n$. As in our linear example, where $Q = \beta(t' - \sum_{i=1}^n T_i)$, we assume that $Q$ is always fully determined by ($\V$, $\F$, $\U$). Given data, the prior $p(\V, \F, \U)$ and the causal query $Q$ induces a posterior density over causal effects, $p(Q|\V)$, as follows, where $\mathcal{A}$ is the set of all tuples $(\F, \U)$ that induce a causal effect $Q$:

\begin{equation}
    p(Q|\V) = \dfrac{1}{p(\V)}\int_{(\F, \U) \in \mathcal{A}} p(\V|\F, \U) \; dp(\F, \U) \numberthis \label{eq:posterior}
\end{equation}

\paragraph{Priors over Structural Functions.} As it is impossible to define a prior over all functions $f: \mathbb{R} \to \mathbb{R}$~\cite{aumann1961borel}, we restrict our attention to structural functions that are fully specified by a finite collection of parameters, $\theta \in \mathbb{R}^d$, with a corresponding prior, $p(\theta)$. While this appears to be restrictive, this template covers a broad range of models, from linear models to Bayesian neural networks~\cite{neal2012bayesian}. As an example, in Section~\ref{sec:gp_example} we show how SBI can be used to reason about SCMs with Gaussian process priors~\cite{rasmussen2003gaussian}, i.e. distributions over deterministic functions, $Y_i = f(X_i) = \phi(X_i)^t \W$, $\W \sim \mathcal{N}(\mu, \Sigma)$, where $\phi$ is some basis function.\footnote{While $p(\Y|\X)$ can often be tractably evaluated for Gaussian process models even as $d\to \infty$ by using the \textit{kernel trick}~\cite{rasmussen2003gaussian}, in this work we consider Gaussian processes with a finite collection of basis functions.}

\section{Identifiability in Bayesian Causal Inference}
\label{sec:Bayesian}

In this work we are interested in understanding the key asymptotic properties of $p(Q| \V)$, namely whether posterior mass concentrates around the true causal effect assymptotically. In other words, can the causal effect be identified from data? We define $\eta$-identifiability in this setting as follows:

\begin{definition}
    \label{def:ID}
    $\mathbf{\eta}$\textbf{-identifiability.} Let $(\tilde{\F}, \tilde{\U}) $ be a set of structural functions and latent confounders in the support of the prior, $p(\F, \U)$. Then, a causal query, $Q$, is $\eta$-identifiable given $(\tilde{\F}, \tilde{\U})$ if for a dataset of $n$ instances, $\tilde{\V} \sim p(\V|\tilde{\F}, \tilde{\U})$, $\Prob(|\tilde{Q} - Q| \leq \eta|\tilde{\V}) \to 1$ for some $\eta \in \mathcal{R^+}$ almost surely as $n \to \infty$, where $\tilde{Q}$ is the causal effect induced by $(\tilde{\F}, \tilde{\U}, \tilde{\V})$.\footnote{Importantly, $p(\tilde{Q}|\tilde{\V})$ marginalizes over $(\F, \U)$, and does not condition on the \enquote{known} $(\tilde{\F}, \tilde{\U})$.}
\end{definition}

Even though Definition~\ref{def:ID} is given in terms of an intractable posterior distribution, determining whether a causal effect is $\eta$-identifiable does not require computation or approximation of the posterior directly. Instead, we show that a causal query is $\eta$-identifiable if and only if there do not exist a set of maximum likelihood structural functions and latent confounder in the support of the prior, $(\F', \U')$, that induce causal effects that differ from $(\tilde{\F}, \tilde{\U})$ by more than $\eta$. Proofs of all theorems are provided in the supplementary materials.

\begin{theorem}
    \label{thm:identifiable}
     $Q$ is $\eta$-identifiable given $(\tilde{\F}, \tilde{\U})$ if and only if for a dataset of $n$ instances, $\tilde{\V} \sim p(\V|\tilde{\F}, \tilde{\U})$, there does not exist an $(\F', \U')$ such that $p(\tilde{\V}| \F', \U') =  p(\tilde{\V}| \tilde{\F}, \tilde{\U})$, $|\tilde{Q} - Q'| > \eta$, and $p(\F', \U')/p(\tilde{\F}, \tilde{\U}) > 0$ almost surely as $n \to \infty$. Here, $\tilde{Q}$ and $Q'$ are the causal effects induced by $(\tilde{\F}, \tilde{\U}, \tilde{\V})$ and $(\F', \U', \tilde{\V})$ respectively.
\end{theorem}

Definition~\ref{def:ID} describes identifiability with respect to a single instantiation, $(\tilde{\F}, \tilde{\U})$. Instead, we would like to make statements about whether causal effects can be uniquely identified with high probability across SCMs sampled from the prior. Let $\textrm{ID}(\tilde{\F}, \tilde{\U}, \eta)$ be a function that returns $1$ if $Q$ is $\eta$-identifiable given $(\tilde{\F}, \tilde{\U})$ under Definition~\ref{def:ID}, and $0$ otherwise. Then, we define $(\zeta, \eta)$-identifiability as follows:

\begin{definition}
    \label{def:zID}
    $\mathbf{(\zeta, \eta)}$\textbf{-identifiability.} For some $0 \leq \zeta \leq 1$, $\eta \in \mathbb{R}^+$, a causal query, $Q$, is $(\zeta, \eta)$-identifiable given a prior distribution $p(\F, \U)$ if the probability that $Q$ is $\eta$-identifiable given a $(\tilde{\F}, \tilde{\U}) \sim p(\F, \U)$ is greater than or equal to $\zeta$, i.e. $\zeta \leq \int \textrm{ID}(\tilde{\F}, \tilde{\U}, \eta) \; dp(\tilde{\F}, \tilde{\U})$.\footnote{The standard Definition 3.2.4 in (Pearl, 2009) is equivalent to our Definition~\ref{def:zID} with $\eta = 0, \zeta = 1$.}
\end{definition}

\subsection{Example: Confounded Linear Model}

\label{sec:lin_example}
Here, we illustrate the Bayesian approach with a linear parametric example over observed $\V = \{\T, \Y\}$ and latent $\U = \{ \ul \}$ and $\Eps = \{\eps_t, \eps_y \} $, which is a simplified version of the example in Section 5 of~\cite{d2019multi}. This example corresponds to the graphical structure shown in Figure~\ref{fig:not_id_contour}. Here, the structural causal model is parameterized by $\theta = \{\gamma, \beta, \alpha, \sigma_U^2, \sigma_T^2, \sigma_Y^2\}$. Assume the following SCM:

\begin{align*}
U_i &\sim \mathcal{N}(0, \sigma_{U}^2) & T_i &= \gamma U_i + \epsilon_{T_i} & Y_i &= \beta T_i + \alpha U_i + \epsilon_{Y_i} \\
\epsilon_{T_i} &\sim \mathcal{N}(0, \sigma_{T}^2) & \epsilon_{Y_i} &\sim \mathcal{N}(0, \sigma_{Y}^2)
\end{align*}

Let our causal query again be the sample average treatment effect (SATE), i.e. $Q(\Y, \Y') = \beta(t' - \sum_{i=1}^n T_i)$. In this setting, estimating the causal effect reduces to estimating $\beta$. As shown in~\cite{d2019multi}, for all $\tilde{\V}$ in the support of $p(\V)$ there exists a set of parameters $\Theta$ such that for all $\theta^{(1)}, \theta^{(2)} \in \Theta$, $\beta^{(1)} \neq \beta^{(2)}$ and $p(\tilde{\V}|\theta^{(1)}) = p(\tilde{\V}|\theta^{(2)})$. In summary, the induced linear system of equations relating parameters to the observable covariance between $\T$ and $\Y$ is rank deficient, leading to non-uniqueness of the maximum likelihood solution. This implies that the posterior odds ratio, $p(\beta^{(1)}|\tilde{\V}) / p(\beta^{(2)}|\tilde{\V})$, reduces to the prior odds ratio, $p(\beta^{(1)}) / p(\beta^{(2)})$, regardless of $n$~\cite{witty2020causal}. It follows straightforwardly that given any nondegenerate prior, $p(\theta)$, $Q$ is therefore not $(\zeta, \eta)$-identifiable for any $0 \leq \zeta \leq 1$, $\eta \in \mathbb{R}^+$. This conclusion agrees with known parametric identification results~\cite{pearl2009causality}.

\begin{algorithm*}[t!]
   \caption{Simulation-Based Identifiability (SBI)}
   \label{alg:ID}
\begin{algorithmic}[1]
    % \STATE {\bfseries procedure} $\textrm{SBI}(p(\F, \U), Q, \zeta, \psi, \alpha)$
    \STATE {\bfseries procedure} $\textrm{SBI}(p(\F, \U, \V), Q, \eta, \zeta)$
   \begin{ALC@g}
   \STATE \textbf{parameters:} $m$, SCM samples; $n$; dataset size; $k$, data samples; $\lambda$, repulsion strength, $\alpha$, significance level
   
   \FOR{$i=1$ {\bfseries to} $m$}
   \STATE  $\tilde{\F}_i, \tilde{\U}_i \sim p(\F, \U)$ \;\;\;\;\;\;\;\;\;\;\;\;\;\;\;\;\;\;\;\;\;\;\;\;\;\;\;\;\;\;\;\;\;\;\;$\triangleright$ Sample structural functions and $n$ confounder instances from the prior
   \FOR{$j=1$ {\bfseries to} $k$}
   \STATE $\tilde{\V}_{i,j} \sim p(\V|\tilde{\F}_i, \tilde{\U}_i)$ \;\;\;\;\;\;\;\;\;\;\;\;\;\;\;\;\;\;\;\;\;\;\;\;\;\;\;\;\;\;\;\;\;\;\;\;\;\;\;\;\;\;\;\;\;\;\;\;\;\;\;\;\;\;\;\;\;\;\;\;\;\;\;\;\;\;\;\; $\triangleright$ Sample $n$ observations from the $i$'th SCM
   \STATE $\Delta \hat{Q}_{i, j} \gets \textrm{Optimize } \mathcal{L}(\cdot, \tilde{\V}_{i, j}; \lambda)$  \;\;\;\;\;\;\;\;\;\;\;\;\;\;\;\;\;\;\;\;\;\;\;\;\;\;\;\;\;\;\;\;\;\;\;\;\;$\triangleright$ Using stochastic gradient descent. See Equation~\ref{eq:SBI}
  \ENDFOR
  \STATE $\hat{\mu}_i \gets \sum_{j=1}^k \Delta \hat{Q}_{i, j} / k$ \;\;\;\;\;\;\;\;\;\;\;\;\;\;\;\;\;\;\;\;\;\;\;\;\;\;\;\;\;\;\;\;\;\;\;\;\;\;\;\;\;\;\;\;\;\;\;\;\;\;\;\;\;\;\;\;\;\;\;\;\;\;\;\;\;\;\;\;\;\;\;\;\;\;\;\;$\triangleright$ Compute sample mean for $i$'th SCM \\
  \STATE $\hat{S}_i \gets \sum_{i=1}^k (\hat{\mu}_i - \Delta \hat{Q}_{i, j})^2 / (k-1)$ \;\;\;\;\;\;\;\;\;\;\;\;\;\;\;\;\;\;\;\;\;\;\;\;\;\;\;\;\;\;\;\;\;\;\;\;\;\;\;\;\;\;\;\;\;\;\;\;\;\;\;$\triangleright$ Compute sample variance for $i$'th SCM \\
   \ENDFOR
   \STATE $l_0 \gets \max_{\zeta' \in [0, \zeta]} \sum_{i=1}^m \log (\mathcal{N}(\hat{\mu}_i; \min(\hat{\mu}_i, \eta), \hat{S}_i/k) \zeta' + \mathcal{N}(\hat{\mu}_i; \max(\hat{\mu}_i, \eta), \hat{S}_i/k) (1-\zeta'))$\;\; $\triangleright$ Null log likelihood
   \STATE $l_a \gets \max_{\zeta' \in [0, 1]} \sum_{i=1}^m \log (\mathcal{N}(\hat{\mu}_i; \min(\hat{\mu}_i, \eta), \hat{S}_i/k) \zeta' + \mathcal{N}(\hat{\mu}_i; \max(\hat{\mu}_i, \eta), \hat{S}_i/k) (1-\zeta'))$ \;\;$\triangleright$ Alt. log likelihood
   \STATE \textbf{return} TRUE \algorithmicif\ $\chi^2(2 (l_a - l_0); 1) < \alpha$ \algorithmicelse\ \textbf{return} FALSE\;\;\;\;\;\;\;\;\;\;\;\;\;\;\;\;\;\; $\triangleright$ Chi-squared test of statistical significance
  \end{ALC@g}
  \end{algorithmic}
\end{algorithm*}
\raggedbottom

\section{Simulation-Based Identifiability}
\label{sec:Optimization}
For the linear example in Section~\ref{sec:lin_example}, we were able to relate the observable covariance between $\T$ and $\Y$ to the latent parameters $\theta$ algebraically. However, it is not clear how we might derive similar results for nonlinear structural functions in general. Instead, we propose an approach for determining causal identifiability using a particle-based optimization scheme which we call simulation-based identifiability (SBI). In summary, SBI uses gradient-based search to find two sets of maximum likelihood structural functions and latent confounders in the support of $p(\F, \U)$, $(\F^{(1)}, \U^{(1)})$ and $(\F^{(2)}, \U^{(2)})$, that induce different causal effects, $Q^{(1)}$ and $Q^{(2)}$, respectively. Let $\lambda \in \mathcal{R^+}$ be a hyperparameter and $\Delta Q \coloneqq |Q^{(1)} - Q^{(2)}|$. Then, consider the following objective function:

\begin{align*}
    \mathcal{L}&(\underbrace{\F^{(1)}, \U^{(1)}}_{\textrm{SCM 1}}, \underbrace{\F^{(2)}, \U^{(2)}}_{\textrm{SCM 2}}, \underbrace{\tilde{\V}\vphantom{g}}_{\textrm{Data}}; \lambda) \numberthis \label{eq:SBI}\\
    &= \underbrace{\log p(\tilde{\V}|\F^{(1)}, \U^{(1)})}_{\textrm{SCM 1 log likelihood}} + \underbrace{\log p(\tilde{\V}|\F^{(2)}, \U^{(2)})}_{\textrm{SCM 2 log likelihood}} + \underbrace{\lambda\Delta Q}_{\textrm{Repulsion}}
\end{align*}

Let $\hat{\F}^{(1)}, \hat{\U}^{(1)}, \hat{\F}^{(2)}, \hat{\U}^{(2)}$ denote a solution that maximizes $\mathcal{L}$, and let $\Delta \hat{Q}$ be the corresponding optimal $\Delta Q$. The following asymptotic theorems hold for any $\lambda \in \mathbb{R}^+$ and bounded $Q$:

\begin{theorem}
    \label{thm:main}
    A causal query $Q$ is $\eta$-identifiable given $(\tilde{\F}, \tilde{\U})$ for a dataset of $n$ instances, $\tilde{\V} \sim p(\V|\tilde{\F}, \tilde{\U})$, if $\Delta \hat{Q} \leq 2 \eta$ and only if $\Delta \hat{Q} \leq \eta$ almost surely as $n \to \infty$. 
\end{theorem}

\begin{theorem}
    \label{thm:p-main}
    A causal query $Q$ is $(\zeta, \eta)$-identifiable given a prior $p(\F, \U)$ for $m$ samples of functions and confounders, $\tilde{\F}_i, \tilde{\U}_i \sim p(\F, \U)$, and $m$ datasets of $n$ instances, $\tilde{\V}_i \sim p(\V|\tilde{\F}_i, \tilde{\U}_i)$, if $\zeta < \sum_{i=1}^m \1_{\Delta \hat{Q}_i > 2\eta}$ and only if $\zeta < \sum_{i=1}^m \1_{\Delta \hat{Q}_i > \eta}$ almost surely as $n, m \to \infty$.
\end{theorem}

Theorems~\ref{thm:main} and \ref{thm:p-main} provide necessary and sufficient conditions for determining identifiability in the limit of infinite simulations given exact solutions to $\mathcal{L}$. However, given finite $n$ and $m$ and approximate solutions to $\mathcal{L}$, $\Delta \hat{Q}$ may be large even if the query is identifiable. To address the problem of finite $n$ and $m$ we propose a likelihood ratio hypothesis test using gradient-based approximate solutions to $\mathcal{L}$. The details of this procedure are shown in Algorithm~\ref{alg:ID}, which works as follows. Repeatedly sample a set of functions and latent confounders, $(\tilde{\F}, \tilde{\U})$, from the prior. For each, repeatedly sample a set of observations, $\tilde{\V}$, and optimize $\mathcal{L}$ jointly for two SCMs, resulting in an approximately optimal $\Delta \hat{Q}$ for the simulated data. Then, apply a likelihood ratio test to determine if the distance between particles is statistically significantly greater than $\eta$ with probability $\zeta$. 

\begin{theorem}
    \label{thm:hyp_alg}
    For convex $\mathcal{L}$, Algorithm~\ref{alg:ID} approaches the most powerful exact test with significance $\alpha$ as $n, k \to \infty$.
\end{theorem}

For finite $k$, where the central limit theorem does not provide an exact description of the distribution of the sample mean $\hat{\mu}_i$, this procedure is best described as an approximate test. See the supplementary materials for additional details about the likelihood ratio test. While gradient-based optimization is not gauranteed to escape local optima, our many experiments in Section~\ref{sec:Experiments} suggest that SBI is robust even when $\mathcal{L}$ is non-convex and for finite $n$, $m$, and $k$. SBI correctly determines identifiability for all six of our latent variable model benchmarks, which we strongly suspect all have non-convex likelihoods. We believe that approximate solutions to $\mathcal{L}$ are reliable in practice for two reasons. First, SBI aggregates $m \cdot k$ independent runs of gradient-based optimization on simulated data in its statistical test. For example, even though 14 of the 5000 trajectories had $\Delta \hat{Q} > \eta$, SBI concluded that SATE for the linear IV benchmark is identifiable. Second, SBI uses stochastic gradients and modern optimizers (e.g., Adam) that are known to escape local optima in non-convex high-dimensional settings.

\paragraph{Selecting the repulsion strength, $\lambda$.} While the choice of repulsion strength, $\lambda$, does not influence our asymptotic results, this is not generally the case for any finite $n$. In our experiments in Section~\ref{sec:Experiments}, we find that even small values of $\lambda$ produce large $\Delta \hat{Q}$ for non-identifiable models.

\subsection{Example: Confounded Gaussian Process}
\label{sec:gp_example}

Let us again consider the confounded model in Section~\ref{sec:lin_example}, instead assuming that the function $Y_i = f(T_i, U_i, \epsilon_{Y_i})$ is drawn from the following Gaussian process prior over $Y_i = \mu_Y(T_i, U_i) + \sigma^2_Y \epsilon_{Y_i}$, where $D \in \mathbb{N}$, $\boldsymbol{\mu}_Y = [\mu_Y(T_1, U_1),..., \mu_Y(T_n, U_n)]$, and $\W = \{\sigma^2_Y, w_0, w_{1, 1}, ..., w_{4, 1}, ..., w_{1, D}, ..., w_{4, D}\}$:

\begin{figure*}[t!]
    \begin{subfigure}[b]{0.32\textwidth}
    \centering
    \includegraphics[width=\columnwidth]{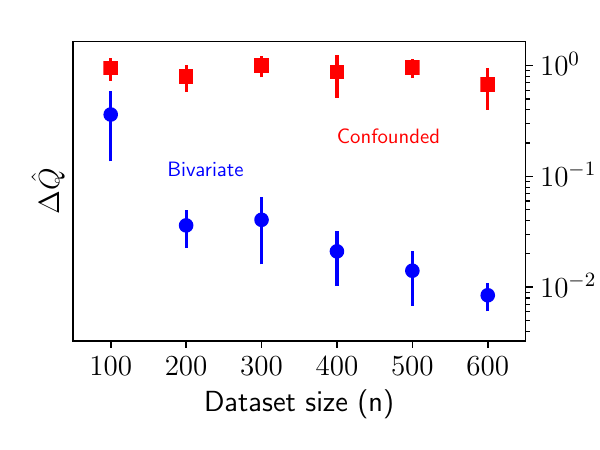}
    \caption{Gaussian Process}
    \label{fig:conf_gp}
    \end{subfigure}
    ~
    \begin{subfigure}[b]{0.32\textwidth}
    \centering
    \includegraphics[width=\columnwidth]{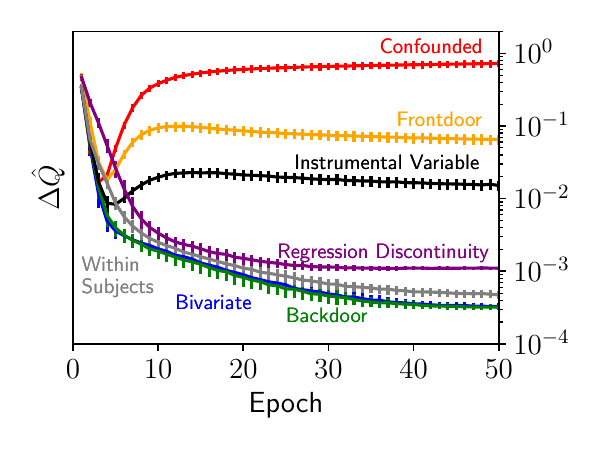}
    \label{fig:linear_training}
    \vspace{-0.5cm}
    \caption{Linear Training Curves}
    \end{subfigure}
    ~
    \begin{subfigure}[b]{0.32\textwidth}
    \centering
    \includegraphics[width=\columnwidth]{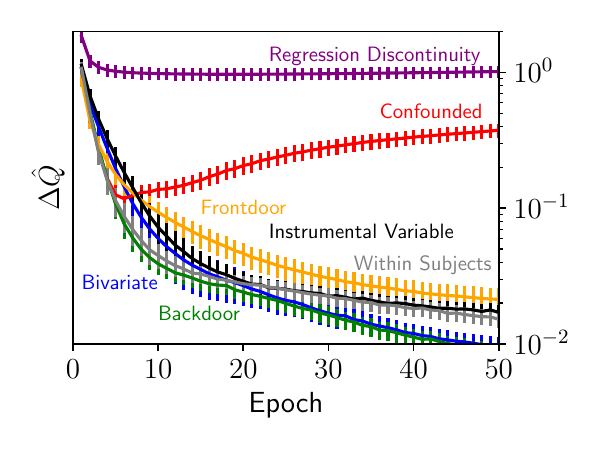}
    \label{fig:gp_training}
    \vspace{-0.5cm}
    \caption{Gaussian Process Training Curves}
    \end{subfigure}
    \caption{\textbf{Summaries of particle-based optimization.} As the simulated dataset size increases the difference between effect estimates of the two particles ($\Delta \hat{Q}$) remains large for the confounded Gaussian process model (a), indicating that the model is not identifiable. Without confounding however, the optimized particles converge to the same causal effect. Using gradient-based optimization, SBI is able to discover likelihood equivalent causal models when they exist that induce different effects for linear (b) and Gaussian process (c) models.}
\end{figure*}

\vspace{-0.5cm}
\begin{align*}
    \mu_Y(T_i, U_i) = w_0 + \sum_{d=1}^D & w_{1, d} \sin(d T_i) + w_{2, d} \cos(d T_i) \\&+ w_{3, d} \sin(d U_i) + w_{4, d} \cos(d U_i)
\end{align*}

This Gaussian process model is known as the Fourier model, where the choice of $D$ and the prior $p(\W)$ dictate the characteristics of the sampled functions~\cite{rasmussen2001occam}. In this and all subsequent experiments we set $D=10$, $w_0 \sim \mathcal{N}(0, 1)$, and $w_{1,d}, w_{2,d}, w_{3,d}, w_{4,d}\overset{iid}\sim \mathcal{N}(0, 1/d^2)$. This choice of prior results in relatively smooth functions, as the weights on higher-order terms are typically close to 0. Again, let the causal query, $Q$, be the sample average treatment effect with the intervention $do(T_i = t')$. Then the log likelihood and the difference between causal effects are given by the following:

\begin{gather*}
\label{eq:GP_likelihood}
\begin{gathered}
    \begin{aligned}
        \log{p(\V|\F, \U)} &= \log{\mathcal{N}(\T; \gamma \ul, \sigma^2_T \I)} + \log{\mathcal{N}(\Y; \boldsymbol{\mu}_Y, \sigma^2_Y \I)}
    \end{aligned}\\
    \begin{aligned}
        \Delta Q &= \sum_{d=1}^D |w_{1, d}^{(1)} - w_{1, d}^{(2)}| \sin(d t') + |w_{2, d}^{(1)} - w_{2, d}^{(2)}| \cos(d t')
    \end{aligned}
\end{gathered}
\end{gather*}

Given this expressions for the log likelihood and the causal query in terms of parameters, $\theta$, and latent confounders, $\U$, we can now compute the partial derivative of the particle-based objective function, $\frac{\dpa}{\dpa s} \mathcal{L} = \frac{\dpa}{\dpa s} \log{p(\tilde{\V}|\F^{(1)}, \U^{(1)})} + \frac{\dpa}{\dpa s} \log{p(\tilde{\V}|\F^{(2)}, \U^{(2)})} + \lambda \frac{\dpa}{\dpa s} \Delta Q$ with respect to all $s \in \theta  \cup \U$. Given an expression for each partial derivative, we can then apply standard gradient-descent algorithms to determine identifiability using SBI. Without loss of generality, the derivative of the repulsion term with respect to $s$ for $\F^{(1)}$, $\U^{(1)}$ is given by the following:

\begin{align*}
    \frac{\dpa}{\dpa s} \Delta Q =   \begin{cases} 
                         \dfrac{w_{1, d}^{(1)} - w_{1, d}^{(2)}}{|w_{1, d}^{(1)} - w_{1, d}^{(2)}|} \sin(dt') & s = w_{1, d}^{(1)}\\
                         \dfrac{w_{2, d}^{(1)} - w_{2, d}^{(2)}}{|w_{1, d}^{(1)} - w_{2, d}^{(2)}|} \cos(dt') & s = w_{2, d}^{(1)}\\
                          0 & \textrm{otherwise}
                      \end{cases}
\end{align*}

For the derivative of the log density we expand on standard identities of Gaussians, where $L_\V$, $L_\T$, and  $L_\Y$ are shorthand for $\log{p(\V|\F^{(1)}, \U^{(1)})}$, $\log{p(\T|\gamma\U, \sigma_T^2\I)}$, and 
$\log{\mathcal{N}(\Y; \boldsymbol{\mu}_Y, \sigma^2_Y \I)}$ respectively:

\begin{align*}
    \frac{\dpa L_\V}{\dpa s}   &= \frac{\dpa L_\T}{\dpa s} + \frac{\dpa L_\Y}{\dpa s}% \\
    % \frac{\dpa L_\T}{\dpa s}  &= \frac{1}{\sigma^2_T} \sum_{i=1}^n (T_i - \gamma U_i)  \frac{\dpa \gamma U_i}{\dpa s} \\
    % & \; \; \; \; \; - \frac{\dpa \sigma^2_T}{\dpa s} \frac{1}{2\sigma_T^2} \bigg(n - \frac{1}{\sigma_T^2}\bigg) \sum_{i=1}^n (T_i - \gamma U_i)^2 \\
    % \frac{\dpa L_\Y}{\dpa s}  &= \frac{1}{\sigma^2_Y} \sum_{i=1}^n (Y_i - \mu_Y(T_i, U_i))  \frac{\dpa \mu_Y(T_i, U_i)}{\dpa s} \\
    % & \; \; \; \; \; - \frac{\dpa \sigma^2_Y}{\dpa s} \frac{1}{2\sigma_Y^2} \bigg(n - \frac{1}{\sigma_Y^2}\bigg) \sum_{i=1}^n (Y_i - \mu_Y(T_i, U_i))^2
\end{align*}

See the supplementary materials for additional details on the remaining partial derivatives. Note that although deriving these gradients is cumbersome and error-prone in general, it can be easily automated using standard automatic differentiation procedures. 

\begin{table*}[t!]
\centering
    \small
    \def\arraystretch{1.1}
  \begin{tabular}{r|c|ccc|c|cccc}
    \toprule 
    Design & Prior
      & $\Delta \hat{Q}_{\textrm{SBI}}$ & $\Delta \hat{Q}_{\textrm{PL}}$ & $\Delta \hat{Q}_{\textrm{MH}}$ & $\textrm{ID}_{\textrm{Truth}}$ & $\textrm{ID}_{\textrm{SBI}}$ & $\textrm{ID}_{\textrm{PL}}$ & $\textrm{ID}_{\textrm{MH}}$ & $\textrm{ID}_{\textrm{DAG}}$\\
     \midrule
    \multirow{2}{*}{Unconfounded} 
    & Linear & .00 $\pm$ .00 & .11 $\pm$ .01 & .12 $\pm$ .03 & \cmark & \cmark & \xmark & \xmark & \cmark\\
    & GP & .01 $\pm$ .00 & .34 $\pm$ .02 & .26 $\pm$ .06 & \cmark & \cmark & \xmark & \xmark & \cmark \\ \hline
    
    \multirow{2}{*}{Confounded} 
    & Linear & .83 $\pm$ .15 & 1.8 $\pm$ .34 & .14 $\pm$ .03 & \xmark & \xmark & \xmark & \xmark & \xmark\\
    & GP & .50 $\pm$ .28 & .73 $\pm$ .14 & .38 $\pm$ .08 & \xmark & \xmark & \xmark & \xmark & \xmark\\ \hline
    
    \multirow{2}{*}{Backdoor} 
    & Linear & .00 $\pm$ .00 & .10 $\pm$ .01 & .11 $\pm$ .02 & \cmark & \cmark & \xmark & \xmark & \cmark\\
    & GP & .01 $\pm$ .00  & .28 $\pm$ .02  & .26 $\pm$ .05 & \cmark & \cmark & \xmark & \xmark & \cmark\\ \hline
    
    \multirow{2}{*}{Frontdoor} 
    & Linear & .06 $\pm$ .04 & .17 $\pm$ .05 & .37 $\pm$ .13 & \cmark & \cmark & \xmark & \xmark & \cmark\\
    & GP & .02 $\pm$ .01 & .20 $\pm$ .09 & .34 $\pm$ .17 & \cmark & \cmark & \xmark & \xmark & \cmark\\ \hline
    
    \multirow{2}{*}{Instrumental variable}  
    & Linear & .01 $\pm$ .01  & .05 $\pm$ .01 & .13 $\pm$ .06 & \cmark & \cmark & \cmark & \xmark & \cmark\\
    & GP & .01 $\pm$ .00  & .37 $\pm$ .03  & .40 $\pm$ .08 & \cmark & \cmark & \xmark & \xmark & \xmark\\ \hline
    
    \multirow{2}{*}{Within subjects} 
    & Linear & .00 $\pm$ .00 & .10 $\pm$ .01 & .14 $\pm$ .03 & \cmark & \cmark & \xmark & \xmark & \xmark\\
    & GP & .01 $\pm$ .01 & .39 $\pm$ .04  & .26 $\pm$ .06 & \cmark & \cmark & \xmark & \xmark & \xmark\\ \hline
    
    \multirow{2}{*}{Regression discontinuity} 
    & Linear & .00 $\pm$ .00  & .16 $\pm$ .01 & .21 $\pm$ .03  & \cmark & \cmark & \xmark & \xmark & \cmark\\
    & GP & 1.1 $\pm$ .12  & 1.1 $\pm$ .09 & .82 $\pm$ .1 & \xmark & \xmark & \xmark & \xmark & \cmark\\ \hline
    \bottomrule
  \end{tabular}
  \caption{\footnotesize{\textbf{Empirical results on quasi-experimental design benchmarks.} Simulation-based identifiability (this paper) correctly determines the identifiability of sample average treatment effects for all fourteen of the benchmark linear and Gaussian process (GP) quasi-experimental designs. Lower $\Delta \hat{Q}$ implies identifiability. The columns labeled $\textrm{ID}$ show whether SBI and the baselines determine the design to be statistically significantly identifiable. Neither of the profile likelihood (PL) or the Metropolis Hastings (MH) baselines consistently determine identifiability. The column labeled $\textrm{ID}_{\textrm{DAG}}$ presents the results of the do-calculus~\cite{pearl1995causal} for GP benchmarks, and IC~\cite{kumor2019efficient} for linear benchmarks applied (incorrectly) to the underlying causal graphs, despite the fact that they do not account for all of the parametric restrictions. This comparison is only to illustrate the effect of parametric restrictions on identifiability.}}
  \label{tab:results}
\end{table*}

Figure~\ref{fig:conf_gp} shows the results of Algorithm~\ref{alg:ID} with this prior over structural causal models using the Adam gradient descent algorithm~\cite{kingma2015adam} to optimize $\mathcal{L}$. Unlike the unconfounded model, which is identical except that $\U$ has been omitted, we conclude that the confounded model is not identifiable. We expand on these examples in Section~\ref{sec:Experiments}.

\section{Experiments}
\label{sec:Experiments}

We evaluated SBI on a benchmark suite of priors reflecting seven standard causal designs which are summarized in Table~\ref{tab:design_desc}; unconfounded regression, confounded regression, backdoor adjusted, frontdoor adjusted, instrumental variable, within-subjects, and regression discontinuity designs. For each of these seven benchmarks we tested SBI using a linear parameterization (e.g. Section~\ref{sec:lin_example}) as well as a parameterization where the outcome function is replaced with a Gaussian process (e.g. Section~\ref{sec:gp_example}). Additional experimental details and descriptions of each prior are provided in the supplementary materials.

We compared SBI against two baselines, one which seeks to approximate the full posterior directly using a Metropolis-Hastings based inference procedure (MH), and one which uses a variation of profile likelihood (PL) identification~\cite{raue2009structural}, which alternates between parameter perturbations and maximum-likelihood optimization. We implemented Algorithm~\ref{alg:ID}, all designs, and the baselines using the Gen probabilistic programming language~\cite{cusumano-towner2019gen}, which provides the necessary support for sampling and automatic differentiation. Using $m=100$, $n=1000$, $k=50$, $\lambda=1$, $\eta = 0.1$, $\zeta = 0.8$, and $\alpha=0.05$, SBI correctly determines the identifiability of all designs, performing significantly better than the two baselines. As we formalized in Section~\ref{sec:Optimization}, if $\Delta \hat{Q}$ is close to $0$ then the causal query is identifiable. 

Our experiments demonstrate that SBI agrees with the do-calculus in settings where graph structure alone is sufficient, and  produces correct identification results for designs that previously required custom identification proofs. Finally, we present the first known identification results for Gaussian process quasi-experimental designs, demonstrating agreement with widely held intuition. See Table~\ref{tab:results} for a summary of SATE identification results. 

\paragraph{Causal graphical models.} In addition to the unconfounded and confounded regression designs presented in Section \ref{sec:gp_example}, we evaluated SBI on two models that are covered by the do-calculus, backdoor-adjusted and frontdoor-adjusted designs. Backdoor-adjusted designs represent settings where all of the random variables that confound the relationship between treatment and outcome are observed, blocking all \textit{backdoor} paths. Unlike backdoor-adjusted designs, frontdoor-adjusted designs can include latent confounding between treatment and outcome, as long as there exists an observed mediator that is not confounded, as in Figure~\ref{fig:id_contour}. Despite this latent confounding, average treatment effects are nonparametrically identifiable~\cite{pearl2009causality}.

\begin{figure*}[t]
    \centering
    \includegraphics[width=\textwidth, trim=0 1.2cm 0 0]{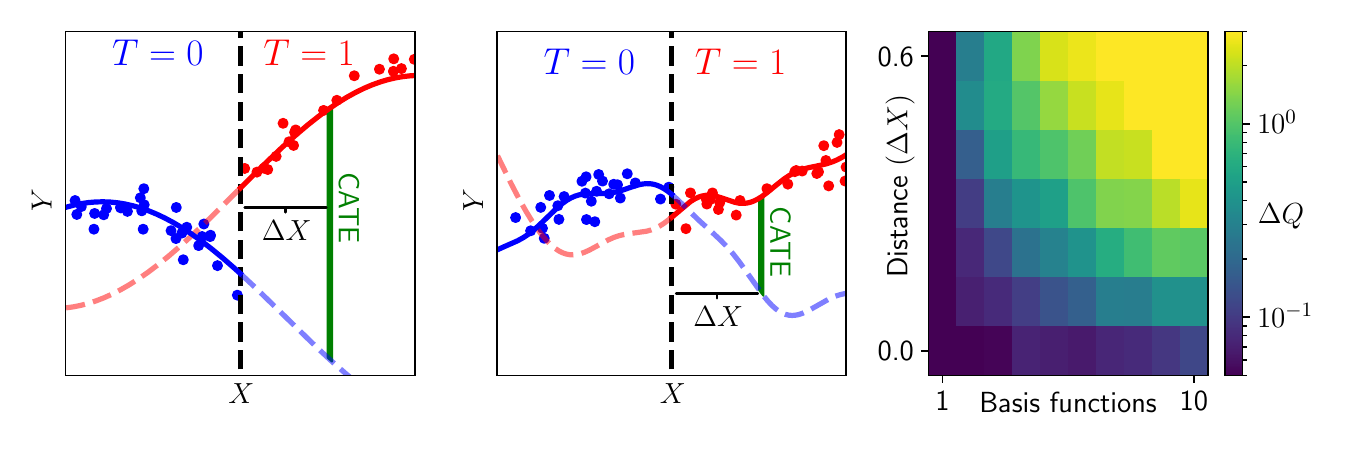}
    \subfloat[\label{fig:rdd_smooth} 1 basis function]{\hspace{0.33\textwidth}}
    \subfloat[\label{fig:rdd_nonsmooth} 10 basis functions ]{\hspace{.33\textwidth}}
    \subfloat[\label{fig:rdd_cate_results} Identifiability heatmap]{\hspace{.33\textwidth}}

    \caption{\textbf{Quantitative insight for conditional average treatment effects.} SBI provides novel and intuitive identification results for the Gaussian process regression discontinuity design benchmark. These results (c) show that conditional average treatment effects (CATE) becomes \textit{less identifiable} as we condition on covariates further from the discontinuity $(\Delta X > 0)$ and for less smooth outcome functions, i.e. increasing the number of basis functions (a, b). }
    \label{fig:RDD}
\end{figure*}
\raggedbottom

% \subsection{Linear Quasi-Experiment Designs}
% \label{sec:lin_qed}
\paragraph{Linear Quasi-Experimental Designs.} Instrumental variable designs differ from the confounded design in that an observed variable, known as the instrument, influences the treatment. Two conditions must be satisfied to enable identification: (i) the instrument and the treatment must not be confounded; and (ii) all influence from the instrument to the outcome is mediated through the treatment. While these assumptions can be expressed graphically, additional parametric assumptions are needed for effects to be identifiable~\cite{pearl2009causality}. For example, if exogenous noise is additive, then the average treatment effect is identifiable~\cite{hartford2017deep}. 

% We evaluated SBI and the baseline on three well-studied, but previously disparate, linear quasi-experimental designs.

Within-subjects designs involve hierarchically structured data in which individual instances (e.g., students) are affiliated with one of several objects (e.g., schools). Treatment effects for these kinds of settings can be identified even if treatment and outcome are confounded, as long as confounders are shared across all instances belonging to the same object~\cite{witty2020causal}. These designs can be described as the family of structural causal models; $T_i = f_T(U_{o(i)}, \epsilon_{T_i})$  $Y_i = f_Y(T_i, U_{o(i)}, \epsilon_{Y_i})$, where $U_{o(i)}$ refers to the shared value of the latent confounder corresponding to instance $i$. Hierarchically structured confounding is applicable to a wide variety of common causal designs~\cite{jensen2020object}: including twin studies~\cite{boomsma2002classical}, difference-in-differences designs~\cite{shadish2008can}, and multi-level-modeling~\cite{gelman2006multilevel}. 

Regression discontinuity designs are quasi-experimental designs in which the treatment depends on a particular observed covariate being above or below a known threshold. We consider a sharp deterministic discontinuity, i.e. $T_i = 1$ if $X_i > 0$, and $T_i = 0$ otherwise. These regression discontinuity designs correspond to the family of structural causal models $X_i = f_X(\epsilon_{X_i})$, $T_i = \1_{X_i > 0}$, and $Y_i = f_Y(T_i, X_i, \epsilon_{T_i})$. Even though all confounders are observed, the deterministic relationship between $X$ and $T$ violates the positivity assumption, which is a necessary assumption for the do-calculus to be sound~\cite{pearl1995causal}. Here, average treatment effects are identifiable for linear models, but not nonparametrically.

\paragraph{Gaussian Process Quasi-Experimental Designs.} We used SBI to determine the previously unknown identifiability of Gaussian process versions of quasi-experimental designs. By assuming a particular kernel we place an inductive bias on the class of structural functions, which could in principle enable identification. SBI instead confirms that the identifibility of these Gaussian process models agrees with the literature on nonparameteric identification.

We also evaluated SBI on the conditional average treatment effect (CATE) for a Gaussian process version of the regression discontinuity design. For nonlinear outcome functions, such as our Gaussian process, observations in one region of $X$ provide only partial information about counterfactuals in another. For example, in Figure~\ref{fig:rdd_nonsmooth} the outcome function for untreated individuals ($T=0$) to the right of the discontinuity (dashed blue curve) is only one of many that are compatible with observed data. Therefore, we should expect that CATE is more ambiguous further from the discontinuity and for less smooth functions. SBI's results in Figure~\ref{fig:rdd_cate_results} agree with this intuition, demonstrating that $\Delta Q$ increases as we condition on covariates further from the discontinuity and as we increase the number of basis functions.

\section{Discussion}
\label{sec:Discussion}

In this paper we demonstrated how SBI can be used to test the identifiability of Bayesian models for causal inference. While determining identifiability is particularly salient in these causal settings, it can also be valuable in non-causal settings as a part of a holistic modeling workflow~\cite{gelman2020bayesian}, supplementing other introspection tools such as simulation-based calbration~\cite{talts2018validating}.

In addition to determining identifiability, SBI can be used as a kind of sensitivity analysis~\cite{franks2019flexible, kallus2019interval, robins2000sensitivity}, bounding the range of causal effects that are likelihood equivalent. Our regression discontinuity design results shown in Figure~\ref{fig:rdd_cate_results} emphasize this capability, showing that irreducible uncertainty in effect estimates increases with increasing distance from the discontinuity and with less smooth outcome functions.

Our benchmarks encode strong parametric assumptions about latent confounders and exogenous noise. If desired, users may represent broader uncertainty using hyperpriors. To demonstrate this, we ran a version of the confounded GP model with additional hyperpriors over the mean and variance of $U$. See the supplementary materials for details. As another example, one could relax additive noise assumptions using Bayesian versions of invertible neural networks~\cite{dinh2016density}, which satisfy SBI's requirements that the likelihood be differentiable and that counterfactual outcomes (and thus $Q$) are fully determined by $(\F, \U, \V)$.

SBI builds on a long history of optimization-focused machine learning research. Reducing identifiability to optimization in this way provides a path towards reasoning about Bayesian models for causal inference at previously unattainable scales. However, this reduction means that SBI's conclusions are dictated by the performance of an approximate global optimization method. Formally quantifying the implications of this approximation error, and extending SBI to discrete combinatorial causal models (e.g. causal discovery) are important areas of future work.

\newpage

\section*{Acknowledgments}
Thanks to Kenta Takatsu, Alex Lew, Cameron Freer, Marco Cusumano-Towner, Tan Zhi-Xuan, Jameson Quinn, Veronica Weiner, Sharan Yalburgi, Przemyslaw Grabowicz, Purva Pruthi, Sankaran Vaidyanathan, Erica Cai, and Jack Kenney for their helpful feedback and suggestions. Sam Witty and David Jensen were supported by DARPA and the United States Air Force under the XAI (Contract No. HR001120C0031), CAML (Contract No. FA8750-17-C-0120), and SAIL-ON (Contract No. w911NF-20-2-0005) programs. Vikash Mansinghka was supported by DARPA under the SD2 program (Contract No. FA8750-17-C-0239), a philanthropic gift from the Aphorism Foundation, a project under the MIT-Takeda program (Proposal No. 51135) and Intel (Agreement No. 6939564). Any opinions, findings and conclusions or recommendations expressed in  this material are those of the authors and do not necessarily reflect the views of DARPA or the United States Air Force. 

\bibliography{ref}

\begin{thebibliography}{53}
\providecommand{\natexlab}[1]{#1}
\providecommand{\url}[1]{\texttt{#1}}
\expandafter\ifx\csname urlstyle\endcsname\relax
  \providecommand{\doi}[1]{doi: #1}\else
  \providecommand{\doi}{doi: \begingroup \urlstyle{rm}\Url}\fi

\bibitem[Angrist et~al.(1996)Angrist, Imbens, and
  Rubin]{angrist1996identification}
Angrist, J.~D., Imbens, G.~W., and Rubin, D.~B.
\newblock Identification of causal effects using instrumental variables.
\newblock \emph{Journal of the American statistical Association}, 91\penalty0
  (434):\penalty0 444--455, 1996.

\bibitem[Athey \& Imbens(2006)Athey and Imbens]{athey2006identification}
Athey, S. and Imbens, G.~W.
\newblock Identification and inference in nonlinear difference-in-differences
  models.
\newblock \emph{Econometrica}, 74\penalty0 (2):\penalty0 431--497, 2006.

\bibitem[Aumann(1961)]{aumann1961borel}
Aumann, R.~J.
\newblock Borel structures for function spaces.
\newblock \emph{Illinois Journal of Mathematics}, 5\penalty0 (4):\penalty0
  614--630, 1961.

\bibitem[Balke \& Pearl(1994)Balke and Pearl]{balke1994counterfactual}
Balke, A. and Pearl, J.
\newblock Counterfactual probabilities: Computational methods, bounds and
  applications.
\newblock In \emph{Uncertainty Proceedings 1994}, pp.\  46--54. Elsevier, 1994.

\bibitem[Bareinboim et~al.(2020)Bareinboim, Correa, Ibeling, and
  Icard]{bareinboim2020pearl}
Bareinboim, E., Correa, J.~D., Ibeling, D., and Icard, T.
\newblock On pearl’s hierarchy and the foundations of causal inference.
\newblock \emph{ACM Special Volume in Honor of Judea Pearl (provisional
  title)}, 2\penalty0 (3):\penalty0 4, 2020.

\bibitem[Bingham et~al.(2018)Bingham, Chen, Jankowiak, Obermeyer, Pradhan,
  Karaletsos, Singh, Szerlip, Horsfall, and Goodman]{bingham2018pyro}
Bingham, E., Chen, J.~P., Jankowiak, M., Obermeyer, F., Pradhan, N.,
  Karaletsos, T., Singh, R., Szerlip, P., Horsfall, P., and Goodman, N.~D.
\newblock {Pyro: Deep Universal Probabilistic Programming}.
\newblock \emph{Journal of Machine Learning Research}, 2018.

\bibitem[Bollen(2005)]{bollen2005structural}
Bollen, K.~A.
\newblock Structural equation models.
\newblock \emph{Encyclopedia of biostatistics}, 7, 2005.

\bibitem[Boomsma et~al.(2002)Boomsma, Busjahn, and
  Peltonen]{boomsma2002classical}
Boomsma, D., Busjahn, A., and Peltonen, L.
\newblock Classical twin studies and beyond.
\newblock \emph{Nature Reviews Genetics}, 3\penalty0 (11):\penalty0 872--882,
  2002.

\bibitem[Branson et~al.(2019)Branson, Rischard, Bornn, and
  Miratrix]{branson2019nonparametric}
Branson, Z., Rischard, M., Bornn, L., and Miratrix, L.~W.
\newblock A nonparametric bayesian methodology for regression discontinuity
  designs.
\newblock \emph{Journal of Statistical Planning and Inference}, 202:\penalty0
  14--30, 2019.

\bibitem[Carpenter et~al.(2017)Carpenter, Gelman, Hoffman, Lee, Goodrich,
  Betancourt, Brubaker, Guo, Li, and Riddell]{carpenter2017stan}
Carpenter, B., Gelman, A., Hoffman, M.~D., Lee, D., Goodrich, B., Betancourt,
  M., Brubaker, M., Guo, J., Li, P., and Riddell, A.
\newblock Stan: A probabilistic programming language.
\newblock \emph{Journal of statistical software}, 76\penalty0 (1), 2017.

\bibitem[Cragg \& Donald(1993)Cragg and Donald]{cragg1993testing}
Cragg, J.~G. and Donald, S.~G.
\newblock Testing identifiability and specification in instrumental variable
  models.
\newblock \emph{Econometric Theory}, pp.\  222--240, 1993.

\bibitem[Cusumano-Towner et~al.(2019)Cusumano-Towner, Saad, Lew, and
  Mansinghka]{cusumano-towner2019gen}
Cusumano-Towner, M.~F., Saad, F.~A., Lew, A.~K., and Mansinghka, V.~K.
\newblock Gen: A general-purpose probabilistic programming system with
  programmable inference.
\newblock In \emph{Proceedings of the 40th ACM SIGPLAN Conference on
  Programming Language Design and Implementation}, PLDI 2019, pp.\  221–236,
  New York, NY, USA, 2019. Association for Computing Machinery.
\newblock ISBN 9781450367127.
\newblock \doi{10.1145/3314221.3314642}.
\newblock URL \url{https://doi.org/10.1145/3314221.3314642}.

\bibitem[D'Amour(2019)]{d2019multi}
D'Amour, A.
\newblock On multi-cause approaches to causal inference with unobserved
  counfounding: Two cautionary failure cases and a promising alternative.
\newblock In \emph{The 22nd International Conference on Artificial Intelligence
  and Statistics}, pp.\  3478--3486, 2019.

\bibitem[Dillon et~al.(2017)Dillon, Langmore, Tran, Brevdo, Vasudevan, Moore,
  Patton, Alemi, Hoffman, and Saurous]{dillon2017tensorflow}
Dillon, J.~V., Langmore, I., Tran, D., Brevdo, E., Vasudevan, S., Moore, D.,
  Patton, B., Alemi, A., Hoffman, M., and Saurous, R.~A.
\newblock Tensorflow distributions.
\newblock \emph{arXiv preprint arXiv:1711.10604}, 2017.

\bibitem[Dinh et~al.(2016)Dinh, Sohl-Dickstein, and Bengio]{dinh2016density}
Dinh, L., Sohl-Dickstein, J., and Bengio, S.
\newblock Density estimation using real nvp.
\newblock \emph{arXiv preprint arXiv:1605.08803}, 2016.

\bibitem[Draper(1995)]{draper1995inference}
Draper, D.
\newblock Inference and hierarchical modeling in the social sciences.
\newblock \emph{Journal of Educational and Behavioral Statistics}, 20\penalty0
  (2):\penalty0 115--147, 1995.

\bibitem[Franks et~al.(2019)Franks, D’Amour, and Feller]{franks2019flexible}
Franks, A., D’Amour, A., and Feller, A.
\newblock Flexible sensitivity analysis for observational studies without
  observable implications.
\newblock \emph{Journal of the American Statistical Association}, 2019.

\bibitem[Gelman(2006)]{gelman2006multilevel}
Gelman, A.
\newblock Multilevel (hierarchical) modeling: What it can and cannot do.
\newblock \emph{Technometrics}, 48\penalty0 (3):\penalty0 432--435, 2006.

\bibitem[Gelman et~al.(2020)Gelman, Vehtari, Simpson, Margossian, Carpenter,
  Yao, Kennedy, Gabry, B{\"u}rkner, and Modr{\'a}k]{gelman2020bayesian}
Gelman, A., Vehtari, A., Simpson, D., Margossian, C.~C., Carpenter, B., Yao,
  Y., Kennedy, L., Gabry, J., B{\"u}rkner, P.-C., and Modr{\'a}k, M.
\newblock Bayesian workflow.
\newblock \emph{arXiv preprint arXiv:2011.01808}, 2020.

\bibitem[Goodman et~al.(2008)Goodman, Mansinghka, Roy, Bonawitz, and
  Tenenbaum]{goodman2008church}
Goodman, N.~D., Mansinghka, V.~K., Roy, D., Bonawitz, K., and Tenenbaum, J.~B.
\newblock Church: a language for generative models.
\newblock In \emph{Proceedings of the Twenty-Fourth Conference on Uncertainty
  in Artificial Intelligence}, pp.\  220--229, 2008.

\bibitem[Hartford et~al.(2017)Hartford, Lewis, Leyton-Brown, and
  Taddy]{hartford2017deep}
Hartford, J., Lewis, G., Leyton-Brown, K., and Taddy, M.
\newblock Deep iv: A flexible approach for counterfactual prediction.
\newblock In \emph{International Conference on Machine Learning}, pp.\
  1414--1423. PMLR, 2017.

\bibitem[Huang \& Valtorta(2006)Huang and Valtorta]{huang2006pearl}
Huang, Y. and Valtorta, M.
\newblock Pearl's calculus of intervention is complete.
\newblock In \emph{Proceedings of the Twenty-Second Conference on Uncertainty
  in Artificial Intelligence}, pp.\  217--224, 2006.

\bibitem[Jensen et~al.(2020)Jensen, Burroni, and Rattigan]{jensen2020object}
Jensen, D., Burroni, J., and Rattigan, M.
\newblock Object conditioning for causal inference.
\newblock In \emph{Uncertainty in Artificial Intelligence}, pp.\  1072--1082.
  PMLR, 2020.

\bibitem[Kallus et~al.(2019)Kallus, Mao, and Zhou]{kallus2019interval}
Kallus, N., Mao, X., and Zhou, A.
\newblock Interval estimation of individual-level causal effects under
  unobserved confounding.
\newblock In \emph{The 22nd International Conference on Artificial Intelligence
  and Statistics}, pp.\  2281--2290. PMLR, 2019.

\bibitem[Kingma \& Ba(2015)Kingma and Ba]{kingma2015adam}
Kingma, D.~P. and Ba, J.
\newblock Adam: A method for stochastic optimization.
\newblock In \emph{ICLR (Poster)}, 2015.

\bibitem[Kumor et~al.(2019)Kumor, Chen, and Bareinboim]{kumor2019efficient}
Kumor, D., Chen, B., and Bareinboim, E.
\newblock Efficient identification in linear structural causal models with
  instrumental cutsets.
\newblock In \emph{Advances in Neural Information Processing Systems}, pp.\
  12477--12486, 2019.

\bibitem[Lee \& Lemieux(2010)Lee and Lemieux]{lee2010regression}
Lee, D.~S. and Lemieux, T.
\newblock Regression discontinuity designs in economics.
\newblock \emph{Journal of economic literature}, 48\penalty0 (2):\penalty0
  281--355, 2010.

\bibitem[Lee et~al.(2020)Lee, Correa, and Bareinboim]{lee2020general}
Lee, S., Correa, J.~D., and Bareinboim, E.
\newblock General identifiability with arbitrary surrogate experiments.
\newblock In \emph{Uncertainty in Artificial Intelligence}, pp.\  389--398.
  PMLR, 2020.

\bibitem[Loftus \& Masson(1994)Loftus and Masson]{loftus1994using}
Loftus, G. and Masson, M.
\newblock Using confidence intervals in within-subject designs.
\newblock \emph{Psychonomic Bulletin \& Review}, 1\penalty0 (4):\penalty0
  476--490, 1994.

\bibitem[Louizos et~al.(2017)Louizos, Shalit, Mooij, Sontag, Zemel, and
  Welling]{louizos2017causal}
Louizos, C., Shalit, U., Mooij, J.~M., Sontag, D., Zemel, R.~S., and Welling,
  M.
\newblock Causal effect inference with deep latent-variable models.
\newblock In \emph{NIPS}, 2017.

\bibitem[Mansinghka et~al.(2014)Mansinghka, Selsam, and
  Perov]{mansinghka2014venture}
Mansinghka, V., Selsam, D., and Perov, Y.
\newblock Venture: a higher-order probabilistic programming platform with
  programmable inference.
\newblock \emph{arXiv preprint arXiv:1404.0099}, 2014.

\bibitem[Neal(2012)]{neal2012bayesian}
Neal, R.~M.
\newblock \emph{Bayesian learning for neural networks}, volume 118.
\newblock Springer Science \& Business Media, 2012.

\bibitem[Neyman \& Pearson(1933)Neyman and Pearson]{neyman1933ix}
Neyman, J. and Pearson, E.~S.
\newblock Ix. on the problem of the most efficient tests of statistical
  hypotheses.
\newblock \emph{Philosophical Transactions of the Royal Society of London.
  Series A, Containing Papers of a Mathematical or Physical Character},
  231\penalty0 (694-706):\penalty0 289--337, 1933.

\bibitem[Pearl(1995)]{pearl1995causal}
Pearl, J.
\newblock Causal diagrams for empirical research.
\newblock \emph{Biometrika}, 82\penalty0 (4):\penalty0 669--688, 1995.

\bibitem[Pearl(2001)]{pearl2001bayesianism}
Pearl, J.
\newblock Bayesianism and causality, or, why i am only a half-bayesian.
\newblock In \emph{Foundations of bayesianism}, pp.\  19--36. Springer, 2001.

\bibitem[Pearl(2009)]{pearl2009causality}
Pearl, J.
\newblock \emph{Causality}.
\newblock Cambridge university press, 2009.

\bibitem[Perov et~al.(2020)Perov, Graham, Gourgoulias, Richens, Lee, Baker, and
  Johri]{perov2020multiverse}
Perov, Y., Graham, L., Gourgoulias, K., Richens, J., Lee, C., Baker, A., and
  Johri, S.
\newblock Multiverse: causal reasoning using importance sampling in
  probabilistic programming.
\newblock In \emph{Symposium on advances in approximate bayesian inference},
  pp.\  1--36. PMLR, 2020.

\bibitem[Rasmussen(2003)]{rasmussen2003gaussian}
Rasmussen, C.
\newblock Gaussian processes in machine learning.
\newblock In \emph{Summer School on Machine Learning}, pp.\  63--71. Springer,
  2003.

\bibitem[Rasmussen \& Ghahramani(2001)Rasmussen and
  Ghahramani]{rasmussen2001occam}
Rasmussen, C.~E. and Ghahramani, Z.
\newblock Occam's razor.
\newblock \emph{Advances in neural information processing systems}, pp.\
  294--300, 2001.

\bibitem[Raue et~al.(2009)Raue, Kreutz, Maiwald, Bachmann, Schilling,
  Klingm{\"u}ller, and Timmer]{raue2009structural}
Raue, A., Kreutz, C., Maiwald, T., Bachmann, J., Schilling, M.,
  Klingm{\"u}ller, U., and Timmer, J.
\newblock Structural and practical identifiability analysis of partially
  observed dynamical models by exploiting the profile likelihood.
\newblock \emph{Bioinformatics}, 25\penalty0 (15):\penalty0 1923--1929, 2009.

\bibitem[Rissanen \& Marttinen(2021)Rissanen and
  Marttinen]{rissanen2021critical}
Rissanen, S. and Marttinen, P.
\newblock A critical look at the consistency of causal estimation with deep
  latent variable models.
\newblock \emph{Advances in Neural Information Processing Systems}, 34, 2021.

\bibitem[Robins et~al.(2000)Robins, Rotnitzky, and
  Scharfstein]{robins2000sensitivity}
Robins, J.~M., Rotnitzky, A., and Scharfstein, D.~O.
\newblock Sensitivity analysis for selection bias and unmeasured confounding in
  missing data and causal inference models.
\newblock In \emph{Statistical models in epidemiology, the environment, and
  clinical trials}, pp.\  1--94. Springer, 2000.

\bibitem[Rubin(1977)]{rubin1977assignment}
Rubin, D.~B.
\newblock Assignment to treatment group on the basis of a covariate.
\newblock \emph{Journal of educational Statistics}, 2\penalty0 (1):\penalty0
  1--26, 1977.

\bibitem[Shadish et~al.(2008)Shadish, Clark, and Steiner]{shadish2008can}
Shadish, W., Clark, M., and Steiner, P.
\newblock Can nonrandomized experiments yield accurate answers? a randomized
  experiment comparing random and nonrandom assignments.
\newblock \emph{Journal of the American Statistical Association}, 103\penalty0
  (484):\penalty0 1334--1344, 2008.

\bibitem[Talts et~al.(2018)Talts, Betancourt, Simpson, Vehtari, and
  Gelman]{talts2018validating}
Talts, S., Betancourt, M., Simpson, D., Vehtari, A., and Gelman, A.
\newblock Validating bayesian inference algorithms with simulation-based
  calibration.
\newblock \emph{arXiv preprint arXiv:1804.06788}, 2018.

\bibitem[Tavares et~al.(2019)Tavares, Zhang, Koppel, and
  Lezama]{tavares2019counterfactual}
Tavares, Z., Zhang, X., Koppel, J., and Lezama, A.~S.
\newblock A language for counterfactual generative models.
\newblock 2019.

\bibitem[Tian \& Pearl(2000)Tian and Pearl]{tian2000probabilities}
Tian, J. and Pearl, J.
\newblock Probabilities of causation: Bounds and identification.
\newblock \emph{Annals of Mathematics and Artificial Intelligence}, 28\penalty0
  (1):\penalty0 287--313, 2000.

\bibitem[Tran \& Blei(2018)Tran and Blei]{tran2018implicit}
Tran, D. and Blei, D.~M.
\newblock Implicit causal models for genome-wide association studies.
\newblock In \emph{International Conference on Learning Representations}, 2018.

\bibitem[Valdes-Sosa et~al.(2011)Valdes-Sosa, Roebroeck, Daunizeau, and
  Friston]{valdes2011effective}
Valdes-Sosa, P.~A., Roebroeck, A., Daunizeau, J., and Friston, K.
\newblock Effective connectivity: influence, causality and biophysical
  modeling.
\newblock \emph{Neuroimage}, 58\penalty0 (2):\penalty0 339--361, 2011.

\bibitem[Witty et~al.(2020{\natexlab{a}})Witty, Lew, Jensen, and
  Mansinghka]{witty2020bayesian}
Witty, S., Lew, A., Jensen, D., and Mansinghka, V.
\newblock Bayesian causal inference via probabilistic program synthesis.
\newblock In \emph{Proceedings of the Second Conference on Probabilistic
  Programming}, 2020{\natexlab{a}}.

\bibitem[Witty et~al.(2020{\natexlab{b}})Witty, Takatsu, Jensen, and
  Mansinghka]{witty2020causal}
Witty, S., Takatsu, K., Jensen, D., and Mansinghka, V.
\newblock Causal inference using gaussian processes with structured latent
  confounders.
\newblock In \emph{International Conference on Machine Learning}, pp.\
  10313--10323. PMLR, 2020{\natexlab{b}}.

\bibitem[Xia et~al.(2021)Xia, Lee, Bengio, and Bareinboim]{xia2021causal}
Xia, K., Lee, K.-Z., Bengio, Y., and Bareinboim, E.
\newblock The causal-neural connection: Expressiveness, learnability, and
  inference.
\newblock 2021.

\bibitem[Zhang et~al.(2021)Zhang, Tian, and Bareinboim]{zhang2021partial}
Zhang, J., Tian, J., and Bareinboim, E.
\newblock Partial counterfactual identification from observational and
  experimental data.
\newblock \emph{arXiv preprint arXiv:2110.05690}, 2021.

\end{thebibliography}
\bibliographystyle{icml2022}

%%%%%%%%%%%%%%%%%%%%%%%%%%%%%%%%%%%%%%%%%%%%%%%%%%%%%%%%%%%%%%%%%%%%%%%%%%%%%%%
%%%%%%%%%%%%%%%%%%%%%%%%%%%%%%%%%%%%%%%%%%%%%%%%%%%%%%%%%%%%%%%%%%%%%%%%%%%%%%%
% APPENDIX
%%%%%%%%%%%%%%%%%%%%%%%%%%%%%%%%%%%%%%%%%%%%%%%%%%%%%%%%%%%%%%%%%%%%%%%%%%%%%%%
%%%%%%%%%%%%%%%%%%%%%%%%%%%%%%%%%%%%%%%%%%%%%%%%%%%%%%%%%%%%%%%%%%%%%%%%%%%%%%%
\newpage
\appendix
\onecolumn
\section{Structural Causal Models}

Here, we present a mathematical description for the structural causal models underlying the unconfounded regression and the backdoor adjusted designs which is agnostic to any particular choice of functions, which we expand on for particular choices of functions in Section~\ref{sec:Supp_Experiments} of this supplementary materials. The remaining five designs in Table 1 are presented throughout the main body of the paper.

\paragraph{Unconfounded Regression.} The unconfounded regression design is identical to the confounded regression design, except that the latent confounded has been omitted. Specifically, we have that $T_i = f_T(\epsilon_{T_i})$ and $Y_i = f_Y(T_i, \epsilon_{Y_i})$.

\paragraph{Backdoor Adjusted Design.} The backdoor adjusted design includes an observed confounder, $X$, that influence both treatment, $T$, and outcome $Y$, but no latent confounders. Specifically, we have that $X_i = f_X(\epsilon_{X_i})$, $T_i = f_T(X_i, \epsilon_{T_i})$, and $Y_i = f_Y(T_i, X_i, \epsilon_{Y_i})$.

\section{Experiments}
\label{sec:Supp_Experiments}

In this section we provide additional detail for the linear and Gaussian process experiments. For each of the seven designs we assume that treatment, $T$, outcome, $Y$, and where applicable covariates, $X$, and instruments, $I$, are observed. All other random variables are latent.

For all of the experiments we used the Adam~\cite{kingma2015adam} algorithm to optimize $\mathcal{L}$. We ran Adam with $\alpha=0.01$, $\beta_1=0.9$, and $\beta_2=0.999$ for fifty epochs with a minibatch size of ten instances. For all of the linear parametric experiments, we assume that each function $V_i = f_V(Pa(V)_i, \epsilon_{V_i}) = \beta \cdot Pa(V)_i + \epsilon_{V_i}$, where each element of $\beta$ is drawn from a normal prior. Here, $Pa(V)_i$ refers to the vector of all latent and observed arguments in the structural function, $f_V$. All of the experiments, including the Gaussian process models, assume that exogenous noise is normally distributed and additive. For all of the Gaussian process experiments, we assume that each outcome function $Y_i = f_Y(Pa(Y), \epsilon_{Y_i})$ is drawn from the same Gaussian process prior described in Section~\ref{sec:gp_example}.

\subsection{Linear Structural Causal Models}

Here, we provide the full prior over linear structural causal models for each of the seven designs in Table 1.

\paragraph{Unconfounded Regression.}

        \begin{align*}
            \beta_Y &\sim \mathcal{N}(1, 0.3) & \log(\sigma_{T}^2) &\sim \mathcal{N}(\negative 1, 0.3) & \log(\sigma_{Y}^2) &\sim \mathcal{N}(\negative 3, 0.3) &
            \epsilon_{T_i} &\overset{iid}\sim \mathcal{N}(0, \sigma_{T}^2) & \epsilon_{Y_i} &\overset{iid}\sim \mathcal{N}(0, \sigma_{Y}^2)\\
            & & T_i &= \epsilon_{T_i} &  Y_i &= \beta_Y \cdot T_i + \epsilon_{Y_i}
        \end{align*}

\paragraph{Confounded Regression.}

        \begin{align*}
            \beta_T &\sim \mathcal{N}(.5, 0.3) & \beta_Y &\sim \mathcal{N}([1, .5]^t, 0.3 \I) & \log(\sigma_{T}^2) &\sim \mathcal{N}(\negative 1, 0.3) & \log(\sigma_{Y}^2) &\sim \mathcal{N}(\negative 3, 0.3) & U_i \overset{iid}\sim \mathcal{N}(0, 0.3)\\
            \epsilon_{T_i} &\overset{iid}\sim \mathcal{N}(0, \sigma_{T}^2) & \epsilon_{Y_i} &\overset{iid}\sim \mathcal{N}(0, \sigma_{Y}^2) & T_i &= \beta_T \cdot U_i + \epsilon_{T_i} &  Y_i &= \beta_Y \cdot [T_i, U_i] + \epsilon_{Y_i}
        \end{align*}

\paragraph{Backdoor Adjusted.}

        \begin{align*}
            \beta_T &\sim \mathcal{N}(.5, 0.3) & \beta_Y &\sim \mathcal{N}([1, .5]^t, 0.3 \I) & \log(\sigma_{X}^2) &\sim \mathcal{N}(\negative 2, 0.3) & \log(\sigma_{T}^2) &\sim \mathcal{N}(\negative 1, 0.3) & \log(\sigma_{Y}^2) &\sim \mathcal{N}(\negative 3, 0.3) \\
            &&\epsilon_{X_i} &\overset{iid}\sim \mathcal{N}(0, \sigma_{X}^2) & \epsilon_{T_i} &\overset{iid}\sim \mathcal{N}(0, \sigma_{T}^2) & \epsilon_{Y_i} &\overset{iid}\sim \mathcal{N}(0, \sigma_{Y}^2)\\ 
            &&X_i &= \epsilon_{X_i} & T_i &= \beta_T \cdot X_i + \epsilon_{T_i} &  Y_i &= \beta_Y \cdot [T_i, X_i] + \epsilon_{Y_i}
        \end{align*}

\paragraph{Frontdoor Adjusted.}

        \begin{align*}
            \beta_T &\sim \mathcal{N}(.5, 0.3) & \beta_X &\sim \mathcal{N}(1, 0.3) & \beta_Y &\sim \mathcal{N}([1, .5]^t, 0.3 \I) & \log(\sigma_{T}^2) &\sim \mathcal{N}(\negative 2, 0.3) & \log(\sigma_{X}^2) &\sim \mathcal{N}(\negative 1, 0.3) \\
            \log(\sigma_{Y}^2) &\sim \mathcal{N}(\negative 3, 0.3) & \epsilon_{T_i} &\overset{iid}\sim \mathcal{N}(0, \sigma_{T}^2) & \epsilon_{X_i} &\overset{iid}\sim \mathcal{N}(0, \sigma_{X}^2) & \epsilon_{Y_i} &\overset{iid}\sim \mathcal{N}(0, \sigma_{Y}^2) & U_i &\overset{iid}\sim \mathcal{N}(0, 0.3) \\ 
            && T_i &= \beta_T \cdot U_i + \epsilon_{T_i} & X_i &= \beta_X \cdot T_i + \epsilon_{X_i} & Y_i &= \beta_Y \cdot [X_i, U_i] + \epsilon_{Y_i}
        \end{align*}

\paragraph{Instrumental Variable.}
        
        Note that here $I_i$ refers to the $i$'th instance of the instrumental random variable $I$, and $\I$ refers to the identity matrix.
        
        \begin{align*}
            \beta_T &\sim \mathcal{N}([2, .5]^t, 0.3 \I) & \beta_Y &\sim \mathcal{N}([1, .5]^t, 0.3 \I) & \log(\sigma_{I}^2) &\sim \mathcal{N}(0, 0.3) & \log(\sigma_{T}^2) &\sim \mathcal{N}(\negative 1, 0.3)  \\
            \log(\sigma_{Y}^2) &\sim \mathcal{N}(\negative 3, 0.3) & \epsilon_{I_i} &\overset{iid}\sim \mathcal{N}(0, \sigma_{I}^2) & \epsilon_{T_i} &\overset{iid}\sim \mathcal{N}(0, \sigma_{T}^2) & \epsilon_{Y_i} &\overset{iid}\sim \mathcal{N}(0, \sigma_{Y}^2) \\ 
            U_i &\overset{iid}\sim \mathcal{N}(0, 0.3) & I_i &= \epsilon_{I_i} & T_i &= \beta_T \cdot [I_i, U_i] + \epsilon_{T_i} & Y_i &= \beta_Y \cdot [T_i, U_i] + \epsilon_{Y_i}
        \end{align*}

\paragraph{Within Subjects.} Here, $U_{o(i)}$ refers to the shared value of the latent confounder, $U_o$, associated with instance $i$. For these and all other experiments, we assume that each object instance, $o$, is shared between $25$ instances of treatment and outcome.
        
        \begin{align*}
            \beta_T &\sim \mathcal{N}(.5, 0.3) & \beta_Y &\sim \mathcal{N}([1, .5]^t, 0.3 \I) & \log(\sigma_{T}^2) &\sim \mathcal{N}(\negative 1, 0.3) & \log(\sigma_{Y}^2) &\sim \mathcal{N}(\negative 3, 0.3) & U_o \overset{iid}\sim \mathcal{N}(0, 0.3)\\
            \epsilon_{T_i} &\overset{iid}\sim \mathcal{N}(0, \sigma_{T}^2) & \epsilon_{Y_i} &\overset{iid}\sim \mathcal{N}(0, \sigma_{Y}^2) & T_i &= \beta_T \cdot U_{o(i)} + \epsilon_{T_i} &  Y_i &= \beta_Y \cdot [T_i, U_{o(i)}] + \epsilon_{Y_i}
        \end{align*}

\paragraph{Regression Discontinuity Design.} Here, $\1[X_i > 0]$ refers to the indicator function that returns $1$ if $X_i > 0$ and $0$ otherwise.
        \begin{align*}
            \beta_Y &\sim \mathcal{N}([0.5, 0.5, \negative 0.5]^t, 0.3\I) & \log(\sigma_{X}^2) &\sim \mathcal{N}(\negative 1, 0.3) & \log(\sigma_{Y}^2) &\sim \mathcal{N}(\negative 3, 0.3) &
            \epsilon_{X_i} &\overset{iid}\sim \mathcal{N}(0, \sigma_{X}^2)  \\
            \epsilon_{Y_i} &\overset{iid}\sim \mathcal{N}(0, \sigma_{Y}^2) & X_i &= \epsilon_{X_i} & T_i &= \1[X_i > 0] & Y_i &= \beta_Y \cdot [X_i, T_i, 1 - T_i] + \epsilon_{Y_i} &
        \end{align*}

\subsection{Gaussian Process Structural Causal Models}

For each of the experiments using Gaussian process priors over structural causal models we use the same prior over linear structural causal models for all functions except the outcome function $f_Y$, which is drawn from the Gaussian process prior described in Section~\ref{sec:gp_example}.
        
\subsection{Baseline}

For our profile likelihood baseline identification method, we used an approach based on profile likelihood identification~\cite{raue2009structural}. For each model the baseline is identical to the SBI in all respects, except that it uses only a single particle with no repulsion term. Instead, to traverse the likelihood surface the baseline first performs $100$ epochs of the Adam optimization method using the gradient of the log-likelihood to find a single maximum likelihood solution. Then, for each parameter $s \in \theta$, we increment the parameter by a small amount $s \gets s + \Delta s$ and then again run the Adam optimization method using the gradient of the log-likelihood with respect to all parameters except for $s$ for $100$ steps. In our experiments we use $\Delta s = 0.01$ for all parameters. We report the range over estimated causal effects after repeating this procedure $100$ times for all $s \in \theta$. Intuitively, if the likelihood surface is on a \textit{ridge} of equivalent maximum likelihood models then alternating between perturbations and optimization will find other locations on that maximum likelihood surface. We discuss limitations of this kind of approach in Section 1, and show empirically that SBI outperforms it in Section 5.

For our Metropolis Hastings baseline identification method, we used a combination of standard inference procedures to approximate the posterior $p(Q|\tilde{V})$ directly. This inference procedure involved alternating between 10 steps of random walk Metropolis Hastings on each $s \in \theta$ and 10 steps of elliptical slice sampling on $U$ (when applicable) a total of $100$ times. To compensate for the additional computational costs of this sampling-based approximate inference procedure, we reduced the number of instances, $(n)$, to $250$ for this baseline. In addition, we eliminated the first $25$ sets of $10$ Metropolis-Hastings and elliptical slice moves as a \textit{burn-in}.

\section{Asymptotic Soundness and Completeness}

In this section, we restate and prove Theorems \ref{thm:identifiable}, \ref{thm:main} and, \ref{thm:p-main}. First, we prove a lemma that the likehood ratio uniformly converges to $0$ or $1$ asymptotically for any pair of SCMs.

\begin{lemma}
    \label{lem:likelihood}
    For all $(\tilde{\F}, \tilde{\U}), (\F', \U')$ in the support of $p(\F, \U)$, $p(\tilde{\V}|\F', \U')/p(\tilde{\V}|\tilde{\F}, \tilde{\U})$ converges uniformly to $0$ or $1$ almost surely as $n \to \infty$, where $\tilde{\V} \sim p(\tilde{\V}|\tilde{\F}, \tilde{\U})$.
\end{lemma}

\begin{proof}
    Let $r_i \coloneqq \mathbb{E}[p(\tilde{\V}_i|\F', \U')/p(\tilde{\V}_i|\tilde{\F},\tilde{\U})] \leq 1$ for a single data instance $i$. As each element of $\epsilon$ is assumed to be independent and identically distributed, then $r_i = r_j = r$ for all $i, j \in 1...n$. Therefore, $\mathbb{E}[p(\tilde{\V}|\F', \U')/p(\tilde{\V}|\tilde{\F},\tilde{\U})] = r^n$ for $n$ i.i.d data instances. As $0 \leq r \leq 1$, $r^n \to 0$ or $1$ uniformly as $n \to \infty$. By the weak law of large numbers, we have that $p(\tilde{\V}|\F', \U')/p(\tilde{\V}|\tilde{\F},\tilde{\U}) \to 0$ or $1$ almost surely for all $(\tilde{\F}, \tilde{\U}), (\F', \U')$ in the support of $p(\F, \U)$ as $n \to \infty$.
\end{proof}

 \textbf{Theorem~\ref{thm:identifiable}.} $Q$ is $\eta$-identifiable given $(\tilde{\F}, \tilde{\U})$ if and only if for a dataset of $n$ instances, $\tilde{\V} \sim p(\V|\tilde{\F}, \tilde{\U})$, there does not exist an $(\F', \U')$ such that $p(\tilde{\V}| \F', \U') =  p(\tilde{\V}| \tilde{\F}, \tilde{\U})$, $|\tilde{Q} - Q'| > \eta$, and $p(\F', \U')/p(\tilde{\F}, \tilde{\U}) > 0$ almost surely as $n \to \infty$. Here, $\tilde{Q}$ and $Q'$ are the causal effects induced by $(\tilde{\F}, \tilde{\U}, \tilde{\V})$ and $(\F', \U', \tilde{\V})$ respectively.

\begin{proof}

% The fact that $Q$ is $\eta$-identifiable if there does not exist such an $(\F', \U')$ follows directly from the Bernstein-von Mises Theorem~\cite{doob1949application}, which states that under certain regularity conditions (e.g. uniqueness of the maximum likelihood) the posterior distribution of any random variable is asymptotically consistent.

Let $\mathcal{A}'$ and $\tilde{\mathcal{A}}$ be the set of $(\F, \U)$ that induce the same effect as $(\F', \U')$ and $(\tilde{\F}, \tilde{\U})$ respectively and let $\mathbb{L}$ be the set of $(\F, \U)$ that maximize the likelihood of the data asymptotically, i.e. $\{(\F, \U) \in \textrm{supp}(p(\F, \U)): \frac{p(\tilde{\V}|\F, \U)}{p(\tilde{\V}|\tilde{\F}, \tilde{\U})} \to 1\ \textrm{ as } n \to \infty\}$.
To show that $Q$ is $\eta$-identifiable only if there does not exist such an $(\F', \U')$, we have that for all $(\F', \U')$ in the support of $p(\F, \U))$:

\begin{align*}
    \lim_{n\to \infty} p(Q'|\tilde{\V}) &= \lim_{n\to \infty} \dfrac{1}{p(\tilde{\V})}\int_{(\F, \U) \in \mathcal{A}'} p(\tilde{\V}|\F, \U) p(\F, \U)  d\F  d\U\\
                         &= \lim_{n\to \infty} \dfrac{p(\tilde{\V}|\tilde{\F}, \tilde{\U})}{p(\tilde{\V})}\int_{(\F, \U) \in \mathcal{A}'} \dfrac{p(\tilde{\V}|\F, \U)}{p(\tilde{\V}|\tilde{\F}, \tilde{\U})} p(\F, \U)  d\F  d\U\\
                         &= \Bigg{(}\lim_{n\to \infty}  \dfrac{p(\tilde{\V}|\tilde{\F}, \tilde{\U})} {p(\tilde{\V})} \Bigg{)} \int_{(\F, \U) \in \mathcal{A}'} \lim_{n\to \infty} \dfrac{p(\tilde{\V}|\F, \U)}{p(\tilde{\V}|\tilde{\F}, \tilde{\U})} p(\F, \U) d\F  d\U\\
                         &= \Bigg{(}\lim_{n\to \infty}  \dfrac{p(\tilde{\V}|\tilde{\F}, \tilde{\U})} {p(\tilde{\V})} \Bigg{)} \int_{(\F, \U) \in \mathcal{A}' \cap \mathbb{L}} \lim_{n\to \infty} p(\F, \U)  d\F  d\U\\
\end{align*}

Here, the limit can be moved inside the integrand by the bounded convergence theorem, as $\frac{p(\tilde{\V}|\F, \U)}{p(\tilde{\V}|\tilde{\F}, \tilde{\U})} p(\F, \U)$ converges uniformly to $p(\F, \U)$ or $0$. Therefore, we have that:

\begin{align*}
    \lim_{n\to \infty} \dfrac{p(Q'|\tilde{\V})}{p(\tilde{Q}|\tilde{\V})} &= \dfrac{\int_{(\F, \U) \in \mathcal{A}' \cap \mathbb{L}} \lim_{n\to \infty} p(\F, \U)  d\F  d\U}{\int_{(\F, \U) \in \tilde{\mathcal{A}} \cap \mathbb{L}} \lim_{n\to \infty} p(\F, \U)  d\F  d\U} > 0 \textrm{ if and only if } \mathcal{A}' \cap \mathbb{L} \neq \emptyset
\end{align*} 

Therefore, if there exists an $(\F', \U') \in \mathcal{A}' \cap \mathbb{L}$ such that $|\Q' - \tilde{\Q}| > \eta$, then $Q$ is not $\eta$-identifiable. If no such $(\F', \U')$ exists, then $Q$ is $\eta$-identifiable. \end{proof}

To prove that SBI is asymptotically sound and complete we first prove that the optimal solutions to $\mathcal{L}$ are almost surely maximum likelihood solutions, and that $\hat{\Q}$ is almost surely the maximum distance between causal effects among maximum likelihood solutions. Recall that $(\hat{\F}^{(1)}, \hat{\U}^{(1)})$ and $(\hat{\F}^{(2)}, \hat{\U}^{(2)})$ are solutions that maximize $\mathcal{L}$.

\begin{lemma}
    \label{lem:like_sol}
    For a dataset of $n$ instances $\tilde{\V} \sim p(\tilde{\V}|\tilde{\F}, \tilde{\U})$, $p(\tilde{\V}|\hat{\F}^{(1)}, \hat{\U}^{(1)})$ and $p(\tilde{\V}|\hat{\F}^{(2)}, \hat{\U}^{(2)})$ converge to $p(\tilde{\V}|\tilde{\F}, \tilde{\U})$ almost surely as $n \to \infty$.
\end{lemma} 

\begin{proof}
    Without loss of generality, toward a contradiction assume that $p(\tilde{\V}|\hat{\F}^{(1)}, \hat{\U}^{(1)}) \not \to p(\tilde{\V}|\tilde{\F}, \tilde{\U})$ as $n \to \infty$. Therefore, by Lemma~\ref{lem:likelihood} we have that $\frac{p(\tilde{\V}|\hat{\F}^{(1)}, \hat{\U}^{(1)})}{p(\tilde{\V}|\tilde{\F}, \tilde{\U})} \to 0$ as $n \to \infty$. $\mathcal{L}(\hat{\F}^{(1)}, \hat{\U}^{(1)}, \hat{\F}^{(2)}, \hat{\U}^{(2)}) \geq \mathcal{L}(\tilde{\F}, \tilde{\U}, \tilde{\F}, \tilde{\U})$ implies that $\log p(\tilde{\V}|\hat{\F}^{(1)}, \hat{\U}^{(1)}) + \log p(\tilde{\V}|\hat{\F}^{(2)}, \hat{\U}^{(2)}) + \lambda |\hat{Q}^{(1)} - \hat{Q}^{(2)}| \geq 2 \log p(\tilde{\V}|\tilde{\F}, \tilde{\U}) + \lambda |\tilde{Q} - \tilde{Q}| = 2 \log p(\tilde{\V}|\tilde{\F}, \tilde{\U})$. Or equivalently, $0 \leq \log p(\tilde{\V}|\hat{\F}^{(1)}, \hat{\U}^{(1)}) + \log p(\tilde{\V}|\hat{\F}^{(2)}, \hat{\U}^{(2)}) + \lambda |\hat{Q}^{(1)} - \hat{Q}^{(2)}| - 2 \log p(\tilde{\V}|\tilde{\F}, \tilde{\U}) = \log \frac{p(\tilde{\V}|\hat{\F}^{(1)}, \hat{\U}^{(1)})}{p(\tilde{\V}|\tilde{\F}, \tilde{\U})} + \log \frac{p(\tilde{\V}|\hat{\F}^{(2)}, \hat{\U}^{(2)})}{p(\tilde{\V}|\tilde{\F}, \tilde{\U})} + \lambda |\hat{Q}^{(1)} - \hat{Q}^{(2)}| \to \log(0) + \log \frac{p(\tilde{\V}|\hat{\F}^{(2)}, \hat{\U}^{(2)})}{p(\tilde{\V}|\tilde{\F}, \tilde{\U})} + \lambda |\hat{Q}^{(1)} - \hat{Q}^{(2)}|  = -\infty$ as $n$ goes to $\infty$, which is a contradiction.
\end{proof}

\begin{lemma}
    \label{lem:max_dist}
    For a dataset of $n$ instances $\tilde{\V} \sim p(\tilde{\V}|\tilde{\F}, \tilde{\U})$, $\Delta \hat{Q} \to \max_{(\F^{(1)}, \U^{(1)}), (\F^{(2)}, \U^{(2)}) \in \mathbb{L}} \Delta Q$ almost surely as $n \to \infty$.
\end{lemma} 

\begin{proof}
    Toward a contradiction assume that there exists some $(\F^{(1)}, \U^{(1)}, \F^{(2)}, \U^{(2)})$  such that $\mathcal{L}(\F^{(1)}, \U^{(1)}, \F^{(2)}, \U^{(2)}) \leq \mathcal{L}(\hat{\F}^{(1)}, \hat{\U}^{(1)}, \hat{\F}^{(2)}, \hat{\U}^{(2)})$ and $|Q^{(1)} - Q^{(2)}| > |\hat{Q}^{(1)} - \hat{Q}^{(2)}|$. By Lemmas~\ref{lem:likelihood} and \ref{lem:like_sol}, we have that as $n \to \infty$, $p(\tilde{\V}|\F^{(1)}, \U^{(1)}) = p(\tilde{\V}|\F^{(2)}, \U^{(2)}) = p(\tilde{\V}|\hat{\F}^{(1)}, \hat{\U}^{(1)}) = p(\tilde{\V}|\hat{\F}^{(2)}, \hat{\U}^{(2)}) = p(\tilde{\V}|\tilde{\F}, \tilde{\U})$. Therefore, by definition of $\mathcal{L}$, we have that $2 \log{p(\tilde{\V}|\tilde{\F}, \tilde{\U})} + |Q^{(1)} - Q^{(2)}| \leq 2 \log{p(\tilde{\V}|\tilde{\F}, \tilde{\U})} + |\hat{Q}^{(1)} - \hat{Q}^{(2)}|$, which implies that $|Q^{(1)} - Q^{(2)}| \leq |\hat{Q}^{(1)} - \hat{Q}^{(2)}|$, which is a contradiction.
\end{proof}

\textbf{Theorem~\ref{thm:main}.} A causal query $Q$ is $\eta$-identifiable given $(\tilde{\F}, \tilde{\U})$ for a dataset of $n$ instances, $\tilde{\V} \sim p(\V|\tilde{\F}, \tilde{\U})$, if $\Delta \hat{Q} \leq 2 \eta$ and only if $\Delta \hat{Q} \leq \eta$ almost surely as $n \to \infty$. 

\begin{proof} 
    By Lemma~\ref{lem:like_sol} we have that $(\hat{\F}^{(1)}, \hat{\U}^{(1)})$ and $(\hat{\F}^{(2)}, \hat{\U}^{(2)})$ are in $\mathbb{L}$, i.e. the set of functions that maximize the log likelihood of the data asymptotically. Therefore, if $|\hat{Q}^{(1)} - \hat{Q}^{(2)}| > 2\eta$, then at least one of $(\hat{\F}^{(1)}, \hat{\U}^{(1)})$ or $(\hat{\F}^{(2)}, \hat{\U}^{(2)})$ are a $(\F', \U')$ that satisfy Theorem~\ref{thm:identifiable}. By Lemma~\ref{lem:max_dist} we have that $|\hat{Q}^{(1)} - \hat{Q}^{(2)}|$ maximizes the distance between induced causal effects. Therefore, if $|\hat{Q}^{(1)} - \hat{Q}^{(2)}| < \eta$ as $n \to \infty$, no such $(\F', \U')$ exists. Note that if $\eta < |\hat{Q}^{(1)} - \hat{Q}^{(2)}| < 2\eta$ we can not conclude whether $\Q$ is $\eta$-identifable, as the true causal effect $\tilde{\Q}$ may be within $\eta$ of either or neither of $\hat{Q}^{(1)}$ or $\hat{Q}^{(2)}$.
\end{proof}

\textbf{Theorem~\ref{thm:p-main}} A causal query $Q$ is $(\zeta, \eta)$-identifiable given a prior $p(\F, \U)$ for $m$ samples of functions and confounders, $\tilde{\F}_i, \tilde{\U}_i \sim p(\F, \U)$, and $m$ datasets of $n$ instances, $\tilde{\V}_i \sim p(\V|\tilde{\F}_i, \tilde{\U}_i)$, if $\zeta < \sum_{i=1}^m \1_{\Delta \hat{Q}_i > 2\eta}$ and only if $\zeta < \sum_{i=1}^m \1_{\Delta \hat{Q}_i > \eta}$ almost surely as $n, m \to \infty$.

\begin{proof}
    Theorem~\ref{thm:p-main} follows directly from the weak law of large numbers applied to the results of Theorem~\ref{thm:main}.
\end{proof}

\section{Likelihood Ratio Hypothesis Test}

Here we provide additional detail for the likelihood ratio test used in Algorithm~\ref{alg:ID}. Recall that $\textrm{ID}(\tilde{\F}, \tilde{\U}, \eta)$ is a function that returns $1$ if $Q$ is $\eta$-identifiable given $(\tilde{\F}, \tilde{\U})$ under Definition~\ref{def:ID}, and $0$ otherwise. Additionally, recall that $\hat{\mu}_i$ is the sample-averaged $\Delta \hat{Q}$ across $k$ datasets drawn from $p(\V|\tilde{\F}_i, \tilde{\U}_i)$ with $n$ instances.

Let $\zeta'$ be the true (unknown) probability that $\textrm{ID}(\tilde{\F}, \tilde{\U}, \eta) = 1$ for $(\tilde{\F}, \tilde{\U}) \sim p(\F, \U)$, let $\textrm{H}_o$ be the null hypothesis that $Q$ is not $(\zeta, \eta)$-identifiable, i.e. $\zeta'  < \zeta$, $\textrm{H}_a$ be the alternative hypothesis that $Q$ is $(\zeta, \eta)$-identifiable, i.e. $\zeta' \geq \zeta$, and let $\textrm{ID}_{\eta, i}$ be shorthand for $\textrm{ID}(\tilde{\F}_i, \tilde{\U}_i, \eta)$.

To construct a likelihood ratio test, we evaluate the maximum of the log data likelihood (here over observed data $\hat{\mu}_i$) in the set of parameters in the null hypothesis, denoted $l_0$, and given the full union of parameters in the null and alternative hypotheses, denoted $l_a$. If the difference between these two quantities is significantly large, i.e. $\chi^2(2 (l_a - l_0); 1) < \alpha$, then we reject the null hypothesis. Intuitively, this test fails to reject the null if adding additional degrees of freedom to the parameter space (here by allowing $\zeta<\zeta'<1$) does not substantially change the maximum of the likelihood.

The following expression gives the maximum of the likelihood for the parameters in the null hypothesis. Here, the likelihood is given with respect to parameters $\theta = \{\zeta', \bar{\mu}_{\textrm{ID}, 1}, ..., \bar{\mu}_{\textrm{ID}, k}, \bar{\mu}_{\textrm{nID}, 1}, ..., \bar{\mu}_{\textrm{nID}, k}\}$. The space of parameters under the null, $\Theta_0$, is defined such that $0 < \zeta' < \zeta$, $\bar{\mu}_{\textrm{ID}, 1}, ..., \bar{\mu}_{\textrm{ID}, k}$ are in $[0, \eta]$, and $\bar{\mu}_{\textrm{nID}, 1}, ..., \bar{\mu}_{\textrm{nID}, k}$ are in $(\eta, \infty)$. Here, $\bar{\mu}_{\textrm{ID}, i}$ and $\bar{\mu}_{\textrm{nID}, i}$ represent the true (unknown) centers for $\hat{Q}$ for the $i$'th SCM when $\textrm{ID}_{\eta, i} = 1$ or $0$ respectively. The space of parameters under the alternative hypothesis, $\Theta_a$, is identical, except that $\zeta < \zeta' < 1$.

\begin{align*}
    l_0 := \max_{\theta \in \Theta_0} \log p(\hat{\mu}_1, ..., \hat{\mu}_k|\theta) &= \max_{\theta \in \Theta_0} \log \prod_{i=1}^m p(\hat{\mu}_i|\theta)\\
    &= \max_{\theta \in \Theta_0} \sum_{i=1}^m \log(p(\hat{\mu}_i|\textrm{ID}_{\eta, i}=1, \theta)p(\textrm{ID}_{\eta, i}=1|\theta) + p(\hat{\mu}_i|\textrm{ID}_{\eta, i}=0, \theta)p(\textrm{ID}_{\eta, i}=0|\theta))\\
    &= \max_{\theta \in \Theta_0} \sum_{i=1}^m \log (\mathcal{N}(\hat{\mu}_i; \bar{\mu}_{\textrm{ID}, i}, \Sigma_i) p(\textrm{ID}_{\eta, i}=1|\zeta') + \mathcal{N}(\hat{\mu}_i; \bar{\mu}_{\textrm{nID}, i}, \Sigma_i) p(\textrm{ID}_{\eta, i}=0|\zeta'))\\
    &= \max_{\zeta' \in [0, \zeta]} \sum_{i=1}^m \log (\mathcal{N}(\hat{\mu}_i; \min(\hat{\mu}_i, \eta), \Sigma_i) \zeta' + \mathcal{N}(\hat{\mu}_i; \max(\hat{\mu}_i, \eta), \Sigma_i) (1-\zeta'))
\end{align*}

Note that the maximum likelihood value of $\bar{\mu}_{\textrm{ID}, i}$ and $\bar{\mu}_{\textrm{ID}, i}$ is given by the closest value to $\hat{\mu}$ in their respective set of possible assignments, resulting in the $\min(\hat{\mu}_i, \eta)$ and $\max(\hat{\mu}_i, \eta), \Sigma_i)$ expressions in the final equation above. By a similar argument, $l_a$ is given by the following expression.

\begin{align*}
    l_a := \max_{\theta \in \Theta_0 \cup \Theta_a} \log p(\hat{\mu}_1, ..., \hat{\mu}_k|\theta) 
    &= \max_{\zeta' \in [0, 1]} \sum_{i=1}^m \log (\mathcal{N}(\hat{\mu}_i; \min(\hat{\mu}_i, \eta), \Sigma_i) \zeta' + \mathcal{N}(\hat{\mu}_i; \max(\hat{\mu}_i, \eta), \Sigma_i) (1-\zeta'))
\end{align*}

\textbf{Theorem~\ref{thm:hyp_alg}.} For convex $\mathcal{L}$, Algorithm~\ref{alg:ID} approaches the most powerful exact test with significance $\alpha$ as $n, k \to \infty$.\\

\begin{proof}
    Theorem~\ref{thm:hyp_alg} follows directly from the Neyman-Pearson lemma~\cite{neyman1933ix} and Theorem~\ref{thm:p-main}.
\end{proof}

\section{Confounded Gaussian Process Kernel Partial Derivatives}

Here, we present a mathematical description of the remaining partial derivatives in Section~\ref{sec:gp_example} with respect to all parameters and latent confounders $s \in \theta \cup \U$.
\begin{align*}
    \frac{\dpa L_\T}{\dpa s}  &= \frac{1}{\sigma^2_T} \sum_{i=1}^n (T_i - \gamma U_i)  \frac{\dpa \gamma U_i}{\dpa s} \\
    & \; \; \; \; \; - \frac{\dpa \sigma^2_T}{\dpa s} \frac{1}{2\sigma_T^2} \bigg(n - \frac{1}{\sigma_T^2}\bigg) \sum_{i=1}^n (T_i - \gamma U_i)^2 \\
    \frac{\dpa L_\Y}{\dpa s}  &= \frac{1}{\sigma^2_Y} \sum_{i=1}^n (Y_i - \mu_Y(T_i, U_i))  \frac{\dpa \mu_Y(T_i, U_i)}{\dpa s} \\
    & \; \; \; \; \; - \frac{\dpa \sigma^2_Y}{\dpa s} \frac{1}{2\sigma_Y^2} \bigg(n - \frac{1}{\sigma_Y^2}\bigg) \sum_{i=1}^n (Y_i - \mu_Y(T_i, U_i))^2
\end{align*}

\begin{align*}
    \frac{\dpa \gamma U_i}{\dpa s} &= \begin{cases} 
                                         U_i & s = \gamma \\
                                         \gamma & s = U_i\\
                                                    0 & \textrm{otherwise}
                      \end{cases}\\
    \frac{\dpa \sigma^2_T}{\dpa s} &= \begin{cases} 
                                         1 & s = \sigma^2_T \\
                                                    0 & \textrm{otherwise}
                      \end{cases}\\   
    \frac{\dpa \sigma^2_Y}{\dpa s} &= \begin{cases} 
                                         1 & s = \sigma^2_Y \\
                                                    0 & \textrm{otherwise}
                      \end{cases}\\
    \frac{\dpa \mu_Y(T_i, U_i)}{\dpa s} &= \begin{cases} 
                                         1 & s = w_0 \\
                                         \sin(d T_i) & s = w_{1, d}\\ 
                                         \cos(d T_i) & s = w_{2, d}\\
                                         \sin(d U_i) & s = w_{3, d}\\ 
                                         \cos(d U_i) & s = w_{4, d}\\
                                         d ( w_{3, d} \cos(d T_i) - w_{4, d} \sin(d T_i))   & s = U_i \\
                                                    0 & \textrm{otherwise}
                      \end{cases}\\ 
\end{align*}

\section{Hyperprior Demonstration}

As we discussed in Section~\ref{sec:Discussion}, we ran an additional experiment to demonstrate the use of hyperpriors to represent broader uncertainty. In this experiment, each $U_i \sim \mathcal{N}(U_\text{mean}, U_\text{var})$, and $U_\text{mean} \sim \mathcal{N}(0, 1)$, $\log(U_\text{var}) \sim \mathcal{N}(0, 1)$. SBI correctly determined that the SATE is not identifable with $\Delta \hat{Q}_{\textrm{SBI}} = 0.55 \pm 0.36$.

\end{document}